\documentclass[runningheads]{llncs}

\usepackage{eccv}

\usepackage{eccvabbrv}

\usepackage{graphicx}
\usepackage{booktabs}
\usepackage[utf8]{inputenc} 
\usepackage[T1]{fontenc}    
\usepackage{url}            
\usepackage{booktabs}       
\usepackage{amsfonts}       
\usepackage{nicefrac}       
\usepackage{microtype}      
\usepackage{xcolor}         
\usepackage{multirow}
\usepackage{subcaption}
\usepackage{graphicx}
\usepackage{enumitem}
\usepackage{comment}
\usepackage{amsmath}
\usepackage{wrapfig}
\def\eg{\emph{e.g.}} 
\def\ie{\emph{i.e.}}

\usepackage{enumitem}
\DeclareMathOperator*{\argmax}{arg\,max}

\usepackage[accsupp]{axessibility}  

\usepackage{hyperref}

\usepackage{orcidlink}

\begin{document}

\title{A Simple Baseline for Spoken Language to \\ Sign Language Translation with 3D Avatars} 

\titlerunning{A Simple Baseline for Spoken2Sign Translation with 3D Avatars}

\author{Ronglai Zuo\inst{1}\thanks{Equal contribution.} \and
Fangyun Wei\inst{2}$^*$\thanks{Corresponding author.} \and
Zenggui Chen\inst{2} \and
Brian Mak\inst{1} \and
Jiaolong Yang\inst{2} \and
Xin Tong\inst{2}
}

\authorrunning{Zuo et al.}

\institute{\textsuperscript{\rm 1}The Hong Kong University of Science and Technology \quad \textsuperscript{\rm 2}Microsoft Research Asia\\
\email{\{rzuo,mak\}@cse.ust.hk} \quad \email{\{fawe,v-zenchen,jiaoyan,xtong\}@microsoft.com}}

\maketitle
\begin{center}
\vspace{-4mm}
\includegraphics[width=\textwidth]{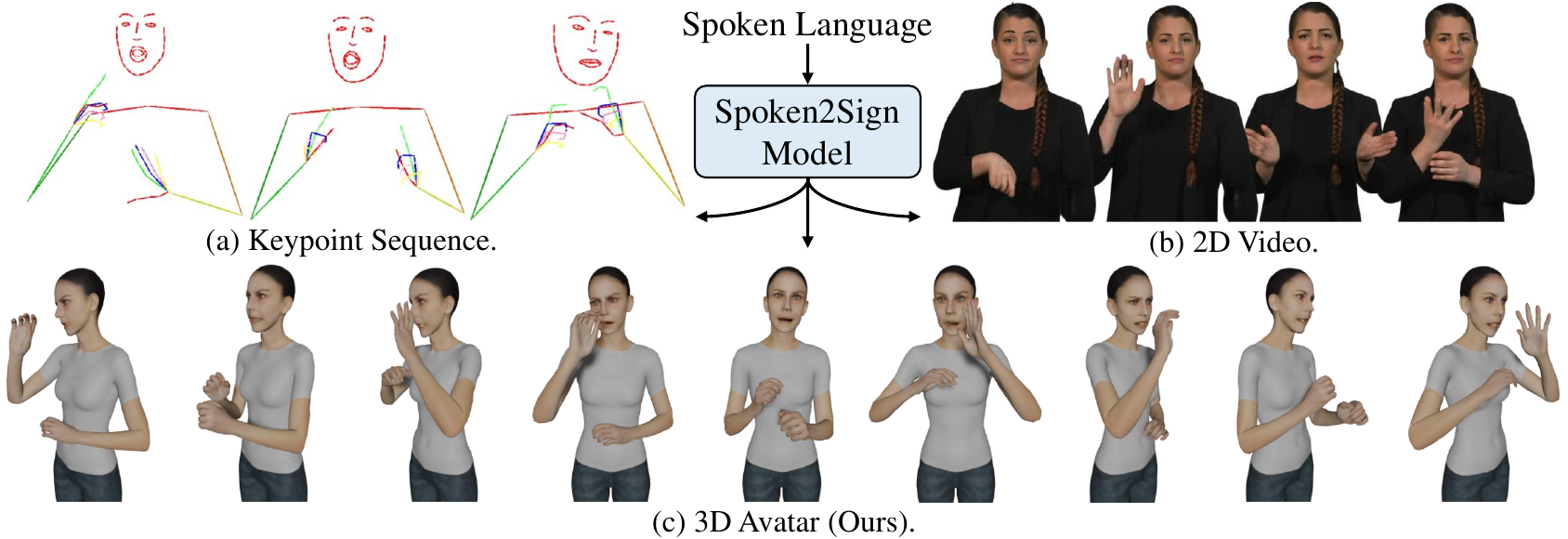}
\vspace{-6mm}
\captionof{figure}{Prior works have presented Spoken2Sign translation results through either (a) keypoint sequences \cite{saunders2021continuous} or (b) 2D videos \cite{saunders2022signing}. In contrast, we utilize a (c) 3D avatar to display the translation results, enabling the visualization of results from any viewpoint.}
\label{fig:teaser}
\end{center}
\vspace{-6mm}

\begin{abstract}
\vspace{-2mm}
The objective of this paper is to develop a functional system for translating spoken languages into sign languages, referred to as Spoken2Sign translation. The Spoken2Sign task is orthogonal and complementary to traditional sign language to spoken language (Sign2Spoken) translation. To enable Spoken2Sign translation, we present a simple baseline consisting of three steps: 1) creating a gloss-video dictionary using existing Sign2Spoken benchmarks; 2) estimating a 3D sign for each sign video in the dictionary; 3) training a Spoken2Sign model, which is composed of a Text2Gloss translator, a sign connector, and a rendering module, with the aid of the yielded gloss-3D sign dictionary. The translation results are then displayed through a sign avatar. As far as we know, we are the first to present the Spoken2Sign task in an output format of 3D signs. In addition to its capability of Spoken2Sign translation, we also demonstrate that two by-products of our approach—3D keypoint augmentation and multi-view understanding—can assist in keypoint-based sign language understanding. Code and models are available at \href{https://github.com/FangyunWei/SLRT}{https://github.com/FangyunWei/SLRT}.
\end{abstract}
\vspace{-9mm}
\section{Introduction}
\vspace{-2mm}
Sign languages are the primary means of communication for the deaf. Numerous previous works~\cite{li2020tspnet,Yao_2023_ICCV,Zhou_2023_ICCV,chentwo,chen2022simple,yin2021simulslt} have focused on sign language translation, with the goal of translating sign languages into spoken languages (Sign2Spoken). However, this paper shifts the focus to the reverse process: translating spoken languages into sign languages (Spoken2Sign) to further bridge the communication gap between the deaf and the hearing.

A majority of prior works~\cite{saunders2020progressive, saundersadversarial, hwang2021non, saunders2021mixed, xie2023vector, huang2021towards, huang2022dualsign, tang2022gloss} on Spoken2Sign translation (aka sign language production) focused on expressing translation outcomes through keypoints (Figure~\ref{fig:teaser}a). However, the keypoint representations often pose interpretable challenges for signers~\cite{slp_review}. With the evolution of generative models \cite{zhang2023adding, bao2017cvae}, several studies \cite{saunders2022signing, fang2023signdiff, stoll2020text2sign} have employed these keypoints to animate signer images, subsequently creating a sign video (Figure~\ref{fig:teaser}b). However, the 2D video format is prone to blurriness and visual distortions. In this work, we introduce an innovative method for Spoken2Sign translation by utilizing a 3D avatar to represent the translation results (Figure~\ref{fig:teaser}c). 
In contrast to earlier attempts that utilized generative models \cite{shmelkov2018good, razavi2019generating}, our method prioritizes enhancing understandability and incorporates a 3D human pose prior with a special emphasis on signing poses, allowing for multi-view representations of the translation results.

We begin with the basic concepts of sign languages:
\begin{itemize}[leftmargin=*]
\setlength\itemsep{-0.5mm}
    \item \textit{Isolated sign.} An isolated sign is a single gesture comprising handshapes, along with movements of the body and hands. Sometimes, facial expressions also play a role in conveying information.
    \item \textit{Continuous sign.} It is a sequence of several isolated signs along with co-articulations.
    \item \textit{Co-articulation.} It refers to the movement of the hands and body between two adjacent signs in a continuous sign.
    \item \textit{Gloss.} A gloss refers to the label assigned to a specific isolated sign. It is usually represented as a word or phrase.
    \item \textit{Gloss sequence.} A label sequence for a continuous sign. In our system, it serves as an intermediate representation.
    \item \textit{Text.} It refers to the translation of a continuous sign. Generally, the \textit{gloss sequence} does not equate with the \textit{text}, as the linguistic rules, \eg, word order, between a sign language and its corresponding spoken language can be different.
\end{itemize}

\begin{figure}[t]
     \centering
     \begin{subfigure}[t]{0.99\textwidth}
         \centering
         \includegraphics[width=\textwidth]{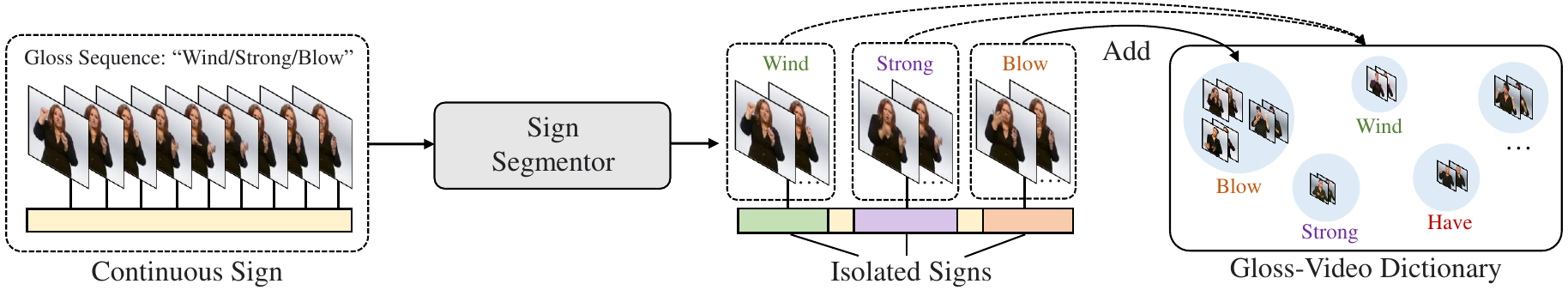}
         \caption{Dictionary construction. We use a well-optimized TwoStream-SLR model~\cite{chentwo} trained on continuous sign language recognition datasets to segment the continuous signs into a set of isolated ones, which are added into the gloss-video dictionary.}
         \label{fig:overview_1}
     \end{subfigure}
     \hfill
     \begin{subfigure}[t]{0.99\textwidth}
         \centering
         \includegraphics[width=\textwidth]{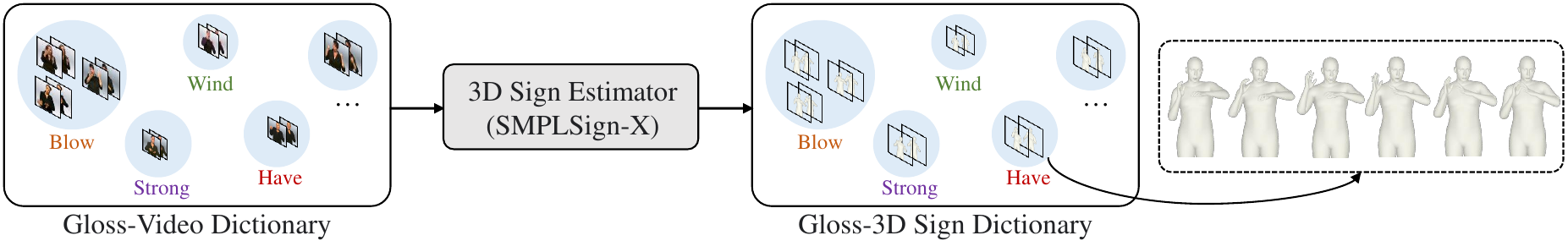}
         \caption{3D sign estimation. We propose SMPLSign-X to estimate the 3D sign for each video in the gloss-video dictionary. A gloss-3D sign dictionary is afterwards acquired.}
         \label{fig:overview_2}
     \end{subfigure}
     \hfill
      \begin{subfigure}[t]{0.99\textwidth}
         \centering
         \includegraphics[width=\textwidth]{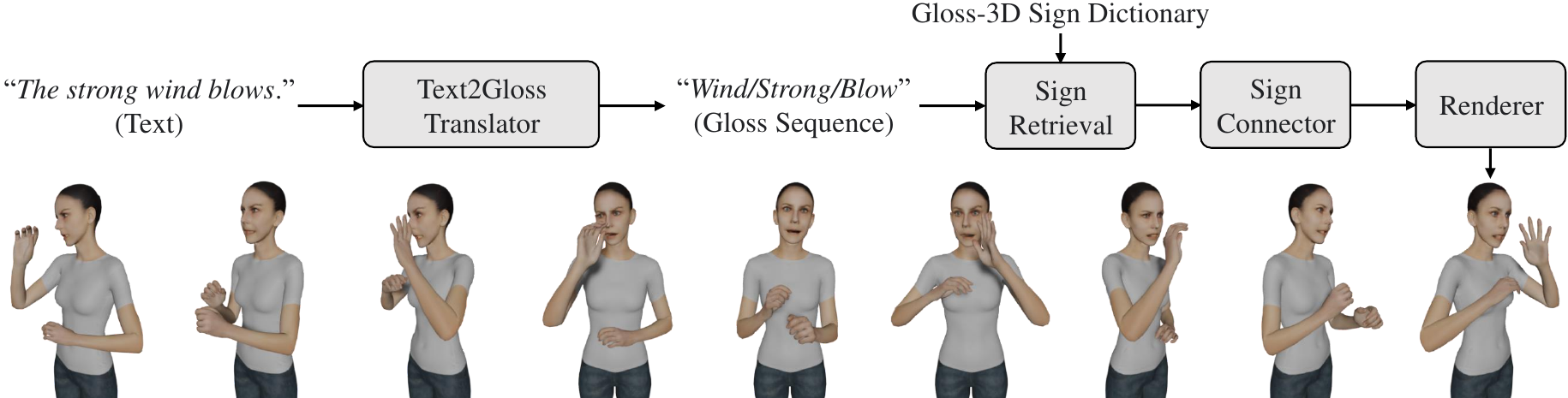}
         \caption{Spoken2Sign translation. We first employ a sequence-to-sequence network mBART~\cite{liu2020multilingual, chentwo} to convert the input text into the intermediate representation, \ie, gloss sequence. We then retrieve the corresponding 3D sign for each gloss in the gloss sequence. A sign connector is trained to generate the co-articulation between two adjacent 3D signs. Finally, we render the output and display the translation results through a 3D avatar.}
         \label{fig:overview_3}
     \end{subfigure}
     \hfill
    \caption{Overview of our methodology. It consists of (a) dictionary construction; (b) 3D sign estimation; (c) Spoken2Sign translation. Sign videos are from Phoenix-2014T~\cite{2014T}, a German sign language benchmark.}
    \label{fig:overview}
    \vspace{-5mm}
\end{figure}

As shown in Figure~\ref{fig:overview}, our Spoken2Sign translation baseline includes three steps: 1) dictionary construction; 2) 3D sign estimation for each entry in the dictionary; and 3) Spoken2Sign translation in a retrieve-then-connect paradigm.

\noindent\textbf{Dictionary Construction.} As shown in Figure~\ref{fig:overview_1}, we initially create a gloss-video dictionary comprising $M$ glosses, each of which may contain multiple isolated sign videos that express the same meaning. Existing sign language recognition (SLR) datasets are typically divided into: 1) datasets for isolated SLR (ISLR), \eg, WLASL~\cite{li2020word} and MSASL~\cite{joze2019ms}; 2) datasets for continuous SLR (CSLR), \eg, Phoenix-2014T~\cite{2014T} and CSL-Daily~\cite{zhou2021improving}. The objective of this work is to develop a Spoken2Sign system, that translates the input text into a sequence of 3D signs, using glosses as intermediate representations. Although existing ISLR datasets are inherent sign language dictionaries, they lack parallel text-gloss sequence data, posing a challenge for the essential text-to-gloss sequence translation. An alternative is to leverage the CSLR datasets; nevertheless, these datasets do not include a dictionary. One solution is utilizing an external dictionary, as demonstrated in FS-Net~\cite{saunders2022signing}. Another approach employs a CSLR model trained with a connectionist temporal classification (CTC) loss~\cite{ctc}, effectively segmenting a continuous sign language video into its constituent signs \cite{Wei_2023_ICCV,zuo2024towards}. The isolated signs produced by this latter method are more suitable for subsequent sign connection since they exclude meaningless motions, such as raising and lowering hands, which may be present in the beginnings and endings of signs in the external dictionary.
In this paper, we adopt the state-of-the-art CTC-based CSLR model, TwoStream-SLR~\cite{chentwo}, as the sign segmentor to construct a dictionary.

\noindent\textbf{3D Sign Estimation.} We currently have a gloss-video dictionary at our disposal. Our next step converts each sign video within this dictionary into a 3D sign. This transformation is essential to mimicking co-articulations more naturally and stitching two signs together in 3D space during Spoken2Sign translation. Inspired by recent advances in 3D whole-body parametric models and monocular 3D whole-body reconstruction~\cite{joo2018total,pavlakos2019expressive,xu2020ghum}, we introduce a sign-language-specific version of SMPLify-X~\cite{pavlakos2019expressive}: SMPLSign-X. This approach is dedicated to estimating 3D signs from monocular sign videos. The vanilla SMPLify-X estimates a holistic and expressive 3D human representation by optimizing shape, pose, and expression parameters. To enhance its application to 3D sign estimation, we incorporate a series of improvements including promoting temporal consistency and involving a more effective 3D pose prior tailored for sign language avatars. Following these improvements, we construct a gloss-3D sign dictionary. The entire process is depicted in Figure~\ref{fig:overview_2}.

\noindent\textbf{Spoken2Sign Translation.} 
Figure~\ref{fig:overview_3} illustrates the core idea behind the proposed Spoken2Sign translation. Glosses serve as a link between sign languages and spoken languages. Therefore, training a text-to-gloss-sequence (Text2Gloss) model is necessary to enable Spoken2Sign translation. Following~\cite{chen2020simple,chentwo}, we adopt mBART~\cite{liu2020multilingual} as our Text2Gloss model due to its promising sequence-to-sequence translation capability. For each gloss in the gloss sequence, we retrieve its corresponding 3D sign from the gloss-3D sign dictionary. A significant challenge arises in seamlessly stitching two adjacent signs, as it is crucial to imitate the co-articulations between them to produce visually appealing translation results. Prior works have either overlooked this challenge \cite{stoll2018sign,zelinka2020neural} or have addressed it through fixed-length interpolation over 2D keypoints \cite{saunders2022signing}.
In this paper, we develop a lightweight sign connector that can dynamically predict the duration of each co-articulation, allowing it to seamlessly stitch adjacent signs in 3D space. 
Our experiments reveal that this simple connector plays a key role in enhancing the overall performance of the Spoken2Sign system.
The final translation results are obtained by rendering the output of the sign connector as a sign avatar.

In summary, the contributions of this work are:
\begin{itemize}
    \item We present a simple yet effective baseline for Spoken2Sign translation. To the best of our knowledge, this is the first work of developing a practical Spoken2Sign system that utilizes a 3D avatar to display the translation results.
    \item Considering the unique characteristics of sign languages, we propose SMPLSign-X, a novel method for 3D sign estimation. Our method aims to develop comprehensive 3D sign language dictionaries for well-established benchmarks including Phoenix-2014T and CSL-Daily. We will release these dictionaries to facilitate future research.
    \item Our Spoken2Sign system sets a new state-of-the-art performance in back-translation evaluation.
    \item Besides demonstrating the capability of Spoken2Sign translation, we also show that two by-products of our approach significantly enhance keypoint-based models for sign language understanding.

\end{itemize}

\vspace{-5mm}
\section{Related Work}
\vspace{-2mm}
\noindent\textbf{Sign Language Understanding.} There are several research directions in sign language understanding, such as sign language recognition (SLR), sign language translation (SLT), sign spotting \cite{albanie2020bsl, momeni2020watch, varol2021read, momeni2022automatic}, and sign language retrieval \cite{duarte2022sign, cheng2023cico}. The objective of SLR is to transcribe a sign video into its constituent glosses. It can be categorized into isolated SLR (ISLR) \cite{li2020word, joze2019ms, hu2021hand, li2020transferring, best, hu2023signbert+, jiang2021skeleton, zuo2023natural} and continuous SLR (CSLR)~\cite{chentwo, zheng2023cvt, vac, self-mutual, stmc, zuo2022c2slr, hu2022self, fcn, chen2022simple, zuo2024improving, niu2024hong}. The former is a classification task aimed at predicting the gloss of an isolated sign, while the latter focuses on recognizing a sequence of signs in a video and generating the corresponding gloss sequence. SLT takes a step further—it aims at translating sign languages into spoken languages. Existing works~\cite{chentwo, chen2022simple, 2014T, camgoz2020multi, slt, li2020tspnet, yin2020better, gan2023contrastive, yu2023efficient} attempt to formulate the SLT task as a neural machine translation problem, using a visual encoder and a language model. Inspired by the recent success of transferring a pre-trained language model to SLT~\cite{chentwo}, we adopt mBART~\cite{liu2020multilingual} as our Text2Gloss translator.

\noindent\textbf{Spoken2Sign Translation.}
In contrast to sign language understanding, this area remains under-explored. Most previous works~\cite{saunders2020progressive, saundersadversarial, hwang2021non, hu2023continuous, saunders2021mixed, xie2023vector, huang2021towards, huang2022dualsign, tang2022gloss} focus on expressing translation results through keypoints, which are often difficult for signers to understand~\cite{slp_review}. With the advances of generative models, several studies \cite{saunders2022signing, fang2023signdiff, stoll2020text2sign, duarte2021how2sign} leverage these keypoints to animate sign videos. However, these approaches often encounter challenges such as blurriness and visual distortions. Some papers~\cite{slp_avatar_1, slp_avatar_2} proposed the use of avatars to display sign languages, but they were too early to benefit from the advanced  parametric 3D human models \cite{pavlakos2019expressive}. In contrast, our approach incorporates a 3D signer pose prior, enhancing both the understandability and the temporal consistency of the animations.

FS-Net \cite{saunders2022signing} represents a notable work in the field. Our approach diverges from FS-Net in four key aspects: 1) we are the pioneers in utilizing 3D avatars as outcomes in Spoken2Sign translation; 2) unlike FS-Net, which relies on an external dictionary, our dictionary is built by segmenting continuous sign videos into isolated ones, which are more suitable for the subsequent sign connection; 3) our lightweight sign connector is designed to flexibly predict the lengths of co-articulations and stitch signs in 3D space, avoiding unrealistic fixed-length interpolation over 2D keypoints; 4) we demonstrate that two by-products of 3D sign representations can significantly aid in the understanding of sign languages.

\noindent\textbf{3D Whole-Body Estimation.}
Recent monocular 3D whole-body estimation approaches adopt parametric models such as Frank~\cite{joo2018total} and SMPL-X~\cite{pavlakos2019expressive} for 3D body representation. These approaches can be divided into fitting-based and regression-based methods. The former utilizes optimization algorithms to fit the parametric model to the given 2D observations, such as body keypoints and segmentation masks~\cite{xiang2019monocular,pavlakos2019expressive,lassner2017unite,xu2020ghum}, while the latter directly predicts model parameters without iterative optimization~\cite{kanazawa2018end,xu2019denserac,lin2023osx,cai2023smpler}, although they often rely on precise SMPL-X annotations~\cite{cai2023smpler}. Parametric models have also been applied in sign language research~\cite{hu2021hand, hu2023signbert+, lee2023human, forte2023reconstructing}. However, none of these studies have developed a complete Spoken2Sign pipeline. In this work, we tailor the widely-used SMPL-X parametric model by incorporating sign-language-specific priors for monocular 3D sign estimation and introduce a functional Spoken2Sign system.

\vspace{-4mm}
\section{Methodology}
\vspace{-2mm}
This section details our methodology, which comprises three stages: dictionary construction (Section~\ref{section: dict}), 3D sign estimation (Section~\ref{section: sign estimation}), and Spoken2Sign translation (Section~\ref{sec:spoken2sign}). Additionally, we discuss two by-products derived from 3D signs in Section~\ref{sec:app}. An overview is depicted in Figure~\ref{fig:overview}.

\vspace{-3mm}
\subsection{Dictionary Construction}
\vspace{-1mm}
\label{section: dict}
Our Spoken2Sign system is built upon the availability of a sign dictionary, which consists of gloss-video pairs. Nevertheless, existing datasets~\cite{2014T, zhou2021improving} do not provide such dictionaries. Inspired by the capability of a well-trained CSLR model to segment a continuous sign video into a collection of isolated signs by finding the optimal alignment path through the CTC forced alignment algorithm~\cite{ctc,sak2015learning,dnf,Wei_2023_ICCV}, we adopt the state-of-the-art CSLR model, TwoStream-SLR~\cite{chentwo}, as the sign segmentor. This model is used to build a sign dictionary containing the segmented isolated signs for each dataset, as shown in Figure~\ref{fig:overview_1}. Subsequently, we acquire a sign dictionary containing $M$ glosses. We also generate a set of co-articulations (which are transitions between two adjacent signs, corresponding to the blank class in the CTC loss~\cite{ctc}) to train our sign connector. More details can be found in the supplementary materials.

\vspace{-3mm}
\subsection{3D Sign Estimation}
\vspace{-1mm}
\label{section: sign estimation}
We introduce the process of estimating the 3D representation for each isolated sign (\ie, a monocular video) in the dictionary, as shown in Figure~\ref{fig:overview_2}.

\noindent\textbf{Preliminaries of SMPLify-X and SMPL-X.} SMPLify-X~\cite{pavlakos2019expressive} is a widely used method to estimate a 3D representation of human body pose, hand pose, and facial expression from a single monocular image. The yielded 3D human representation is termed as a SMPL-X model~\cite{pavlakos2019expressive}, which is defined by a series of learnable parameters. These parameters include global orientation $\zeta \in \mathbb{R}^{3}$, body shape $\beta  \in \mathbb{R}^{10}$, facial expression $\psi \in \mathbb{R}^{10}$, and body pose $\theta \in \mathbb{R}^{3N}$, where $N=54$ denotes the number of joints. Note that pose parameters represent the relative axis-angle rotations to the parent joints defined in a kinematics map. Using these parameters, the SMPL-X model could produce a mesh comprising 10,475 vertices and a set of 118 3D joints $\mathcal{D} \in \mathbb{R}^{118\times 3}$. For simplicity, we use $\mathcal{D}_i \in \mathbb{R}^3$ ($1\leq i \leq 118$) to denote the $i$-th joint in $\mathcal{D}$. Note that only 54 out of the 118 joints are associated with the pose parameters $\theta$.

To fit the SMPL-X model to a single monocular image, we first use HRNet \cite{wang2020deep} pre-trained on COCO-Wholebody \cite{jin2020whole}, to estimate the image's 2D keypoints $\mathcal{K} \in \mathbb{R}^{118\times2}$. We employ only a subset of the COCO-Wholebody keypoints for aligning the joints defined by SMPL-X. Subsequently, SMPLify-X~\cite{pavlakos2019expressive} seeks to minimize the following objective function by optimizing $\zeta$, $\beta$, $\psi$ and $\theta$: 
\begin{equation}
\label{eq:smplify-x}
\mathcal{L}= \mathcal{L}_{joint}+\mathcal{L}_{prior}+\mathcal{L}_{penetration},
\end{equation}
where $\mathcal{L}_{joint}$ is the major loss function that minimizes the distance between the 2D keypoints $\mathcal{K}$ and the projected keypoints $P(\mathcal{D})$; $\mathcal{L}_{prior}$ denotes a combination of losses that incorporate prior knowledge of hand pose, facial pose, body shape, and facial expressions, and penalize extreme body states; $\mathcal{L}_{penetration}$ is a regularization term designed to prevent the SMPL-X model from penetrations and self-collisions. In SMPLify-X, $\mathcal{L}_{joint}$ is formulated as:
\begin{equation}
\label{eq:joint}
    \mathcal{L}_{joint} = \frac{1}{|\mathcal{J}|}\sum_{i\in \mathcal{J}  } \gamma_{i} \omega_{i} \ell_{r} (P\left(\mathcal{D} _{i})-\mathcal{ K}_{i}\right),
\end{equation}
where $\mathcal{J}$ represents the set of 118 3D joints; $P(\cdot)$ is a function that projects each 3D joint $\mathcal{D}_i \in \mathbb{R}^3$ from the world coordinate to image coordinate; $\mathcal{K}_i$ is the associated 2D keypoint (pseudo ground truth) of $\mathcal{D}_i$; $\omega_{i}$ is the confidence (yielded by HRNet) of $\mathcal{K}_i$; $\gamma_{i}$ is the pre-defined weight of joint $\mathcal{D}_i$; $\ell_r$ represents a robust Geman-McClure loss function~\cite{geman1987statistical}. More details can be found in~\cite{pavlakos2019expressive}.

\noindent\textbf{SMPLSign-X.} 
We enhance SMPLify-X by adapting its input from a single monocular \emph{image} to a sign \emph{video}. 
Additionally, we consider the unique properties of sign languages to further improve estimation quality. The new estimator is named SMPLSign-X. Our improvements are based on three observations: 1) optimization targets are absent for joints that do not appear in sign videos (\eg, the lower body and a dropping hand); 2) the upper body of a signer remains upright during signing; 3) independently fitting each frame to the SMPL-X model results in temporal inconsistencies and visually unsatisfactory outcomes. To address these issues, we define the objective function for SMPLSign-X as:
\begin{equation}
\label{eq:smplify-x_improved}
\mathcal{L} = \mathcal{L}_{joint}+\mathcal{L}_{prior}+\mathcal{L}_{penetration} + \lambda_1\mathcal{L}_{unseen}  + \lambda_2\mathcal{L}_{upright} + \lambda_3\mathcal{L}_{smooth}.
\end{equation}
$\mathcal{L}_{unseen}$ (Eq.~\ref{eq:unseen}) is a regularization term that pushes the unseen keypoints to approach those of the rest pose. We use confidence scores predicted by the pre-trained HRNet to identify the unseen keypoints. Concretely, for a 2D keypoint $\mathcal{K}_i$ and its confidence score $\omega_i$, we regard $\mathcal{K}_i$ as an unseen keypoint if $\omega_i<\lambda$, where $\lambda$ is a pre-defined threshold, and $\lambda=0.65$ by default. SMPLify-X provides joint mappings between the 2D keypoints $\mathcal{K}$ and the pose parameters $\theta$. Therefore, we can easily identify the set of unseen joints ($\mathcal{J}_{unseen}$) in the $\theta$ space. We use $\hat{\theta}$ to denote the pose parameters of the rest pose, which are frozen during training.
\begin{equation}
\label{eq:unseen}
\mathcal{L}_{unseen} = \sum_{i\in \mathcal{J}_{unseen}  } \ell_r(\theta_{i}-\hat{\theta}_{i} ).
\end{equation}
$\mathcal{L}_{upright}$ (Eq.~\ref{eq:upright}) denotes a regularization term for encouraging an upright posture in the upper body. To accomplish this, we define a keypoint set $\mathcal{J}_{upright}$, including the neck and pelvis keypoints within the context of $\theta$ space. Our goal is to preserve depth consistency across all keypoints in $\mathcal{J}_{upright}$, as specified by the loss in Eq.~\ref{eq:upright}, where $d$ represents depth.
\begin{equation}
\label{eq:upright}
\mathcal{L}_{upright} = \sum_{i,j\in \mathcal{J}_{upright}  } \ell _r(d_{i}-d_{j} ).
\end{equation}
Finally, given the pose parameters $\theta^{pre}_{i}$ of the previous frame, we use Eq.~\ref{eq:smooth} to preserve the temporal consistency for the current frame. $\mathcal{J}_{\theta}$ is the set containing all joints in the $\theta$ space. $\gamma_{i}$ denotes the weight of the $i$-th joint. These joint weights are pre-defined by SMPLify-X~\cite{pavlakos2019expressive}.
\begin{equation}
\label{eq:smooth}
\mathcal{L}_{smooth} =\sum_{i\in \mathcal{J}_{\theta} }\gamma_{i} \ell _r(\theta _{i}-\theta^{pre}_{i}).
\end{equation}

We estimate the 3D representation for each sign video from the gloss-video dictionary using the objective function in Eq.~\ref{eq:smplify-x_improved}, on a frame-by-frame basis. Subsequently, we construct a gloss-3D sign dictionary.

\vspace{-1mm}
\subsection{Spoken2Sign Translation}
\label{sec:spoken2sign}
As shown in Figure~\ref{fig:overview_3}, our Spoken2Sign translation pipeline primarily consists of three components: 1) a Text2Gloss translator that translates the input text into a gloss sequence; 2) a sign connector, which stitches two adjacent 3D signs together; and 3) a rendering module for producing the final animated sign avatar.

\noindent\textbf{Text2Gloss Translator.}
Inspired by the recent success of sign language translation (SLT)~\cite{chentwo}, we adopt mBART~\cite{liu2020multilingual} as our Text2Gloss translator. Existing SLT benchmarks \cite{2014T, zhou2021improving} typically provide annotations of text-gloss-sequence pairs. In the vein of traditional SLT models, which train a Gloss2Text network using gloss sequences as inputs and texts as outputs, we train a Text2Gloss translator by simply reversing the inputs and outputs. Our Text2Gloss translator achieves a promising BLEU-4 score of 29.27/31.88 on the dev set of Phoenix-2014T/CSL-Daily. More details are available in the supplementary materials.

\noindent\textbf{Sign Retrieval.} 
For each gloss predicted by the Text2Gloss translator, we retrieve its corresponding 3D sign from the gloss-3D sign dictionary. Since a single gloss may be associated with multiple 3D signs, we develop a retrieval strategy to identify the optimal one. Specifically, we train an ISLR model~\cite{zuo2023natural} on all instances in the dictionary. During retrieval, we feed candidate signs into the ISLR model and select the sign with the highest confidence for the gloss query.

\begin{wrapfigure}{r}{0.5\textwidth}
\vspace{-5mm}
  \begin{center}
    \includegraphics[width=0.5\textwidth]{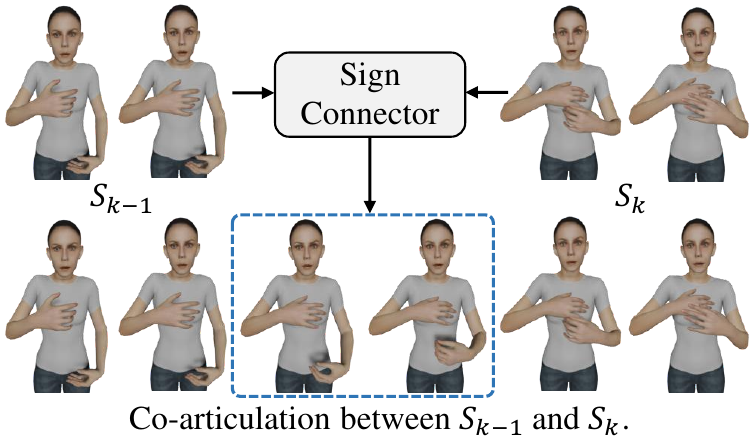}
  \end{center}
  \vspace{-5mm}
  \caption{Illustration of the sign connector. The objective is to predict the duration of the co-articulation between two adjacent 3D signs, $S_{k-1}$ and $S_k$, followed by generating the co-articulation through interpolation in the 3D joint space.}
  \vspace{-2mm}
  \label{fig:conn}
\end{wrapfigure}

\noindent\textbf{Sign Connector.} 
As shown in Figure~\ref{fig:conn}, we present a sign connector to predict the duration of co-articulation between two adjacent 3D signs when they are stitched together. As described in Section~\ref{section: dict}, we generate a set of co-articulations, each of which is denoted as a triplet $(\mathcal{D}^{pre}_{\mathcal{J}_{SC}}, L, \mathcal{D}^{next}_{\mathcal{J}_{SC}})$, where $L$ is the duration of the co-articulation, $\mathcal{J}_{SC}$ denotes the set of 3D joints used in our sign connector, and $\mathcal{D}^{pre}_{\mathcal{J}_{SC}}$ and $\mathcal{D}^{next}_{\mathcal{J}_{SC}}$ represent the 3D joints of the preceding and succeeding signs of the co-articulation, respectively. In our implementation, $\mathcal{J}_{SC}$ includes the joints of hands, wrists, and elbows. Our sign connector is a 4-layer MLP taking the concatenation of $\mathcal{D}^{pre}_{\mathcal{J}_{SC}}$, $\mathcal{D}^{next}_{\mathcal{J}_{SC}}$, and their Euclidean coordinate distance $\mathcal{D}^{pre}_{\mathcal{J}_{SC}} - \mathcal{D}^{next}_{\mathcal{J}_{SC}}$ as inputs. We minimize the loss between the prediction yielded by the MLP and the target $L$ using the L1 loss function. 

Once the sign connector is optimized, given a 3D sign sequence $\{S_1,S_2,...,S_K\}$, it can predict the duration $\hat{L}_k$ of the co-articulation between two adjacent 3D signs $S_{k-1}$ and $S_k$ ($2\leq k \leq K$). Subsequently, we interpolate $\hat{L}_k$ frames between the last frame of $S_{k-1}$ and the first frame of $S_k$ in the 3D joint space to simulate the co-articulation $C^{k-1}_k$. Finally, a 3D sign sequence $\{S_1,C^1_2,S_2,...,S_{K-1},C^{K-1}_K,S_K\}$ is yielded for rendering.

\noindent\textbf{Rendering Module.} 
We render the 3D sign sequence frame by frame using the Blender toolkit \cite{blender}. The pose and facial expression parameters of the SMPL-X model are utilized to drive the avatar.

\subsection{By-Products of 3D Signs}
\label{sec:app}
The generated 3D signs implicitly integrate the human pose prior within the SMPLSign-X. Below, we discuss two by-products of these 3D signs, which significantly enhance keypoint-based models for sign language understanding.

\noindent\textbf{3D Keypoint Augmentation.}
In contrast to existing Spoken2Sign translation works~\cite{saunders2020progressive, saunders2021continuous, saunders2021mixed, saunders2022signing, saundersadversarial}, which only generate frontal-view keypoints/videos, our approach outputs a sequence of 3D signs. The 3D nature of these signs allows the development of 3D keypoint augmentation to enhance keypoint-based models for sign language understanding. Specifically, during each training iteration, we sample an angle, $\delta$, from a pre-defined range of $[-\Delta, \Delta]$, where $\Delta=20^\circ$ by default. We then rotate the original 3D keypoints by $\delta$, modifying the global orientation parameters $\zeta$. Finally, the projected 2D keypoints are fed into the sign language understanding models as inputs.

\noindent\textbf{Understanding from Multiple Views.}
Current datasets for sign language understanding commonly include 2D videos of signs captured from a frontal perspective. This limitation leads to the development of models that are only capable of interpreting sign languages from this single viewpoint. The introduction of 3D signs naturally enables sign language understanding from multiple views.  Drawing inspiration from the TwoStream Network~\cite{chentwo}, which simultaneously processes 2D sign videos and their corresponding keypoint sequences, we modify the network to process two keypoint sequences: one sequence represents the original frontal view (a projection of the 3D keypoints without rotation), while the other sequence represents the side view (a projection of the 3D keypoints resulted from a 60$^\circ$ rotation).

\begin{figure}[!th]
     \centering
     \begin{subfigure}[t]{0.99\textwidth}
         \centering
         \includegraphics[width=\textwidth]{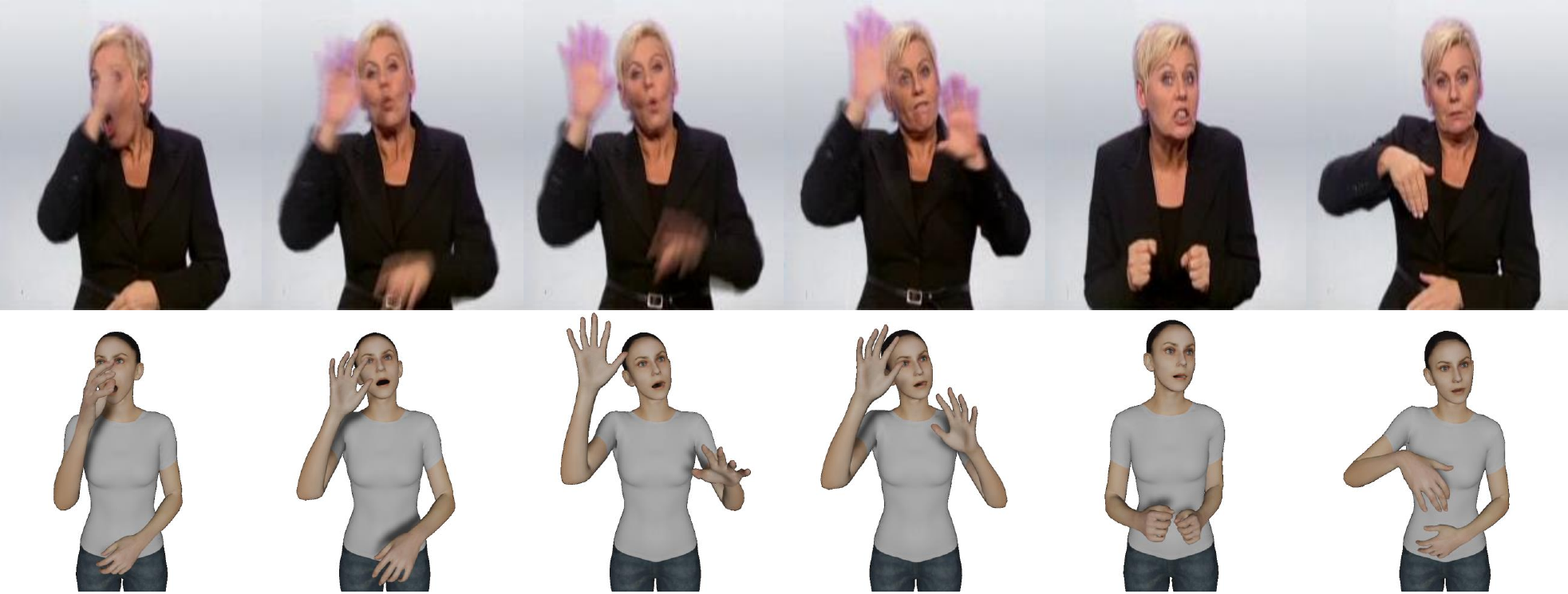}
         \vspace{-4mm}
         \caption{``Clouds and frost exist in the north.''}
         \label{fig:vis_A}
     \end{subfigure}
    \hfill
    \begin{subfigure}[t]{0.99\textwidth}
         \centering
         \includegraphics[width=\textwidth]{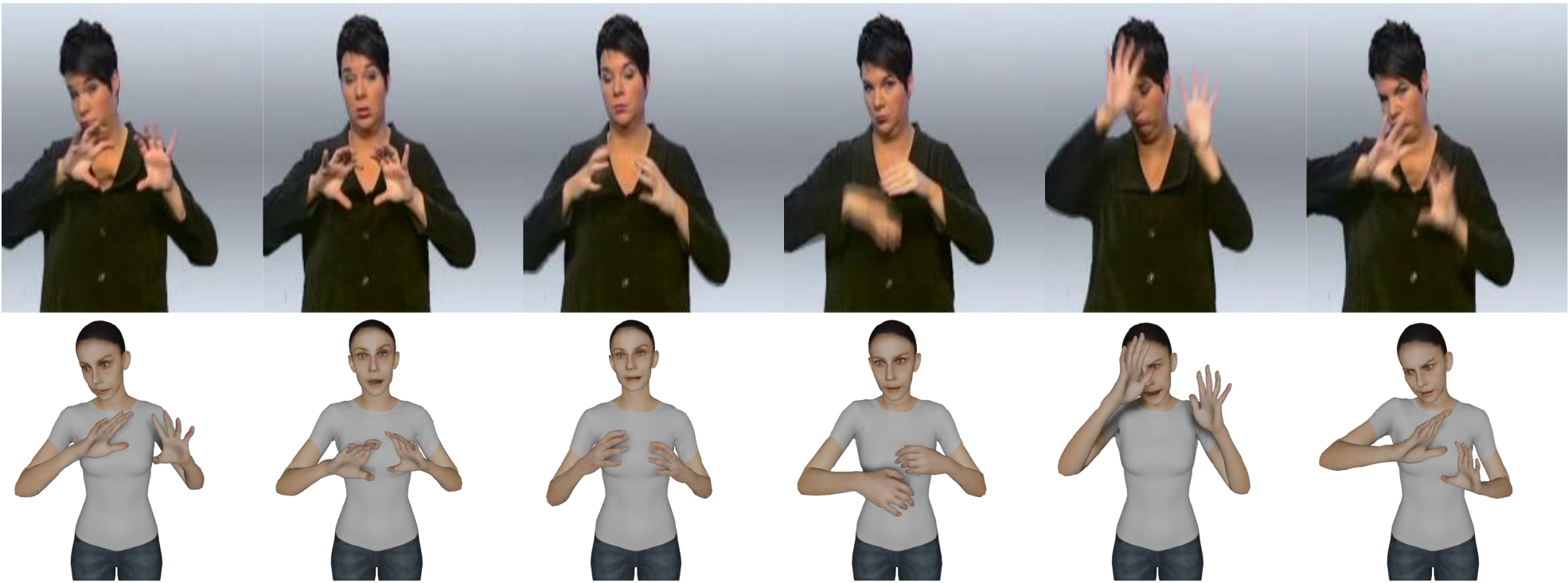}
         \vspace{-4mm}
         \caption{``There are more rain and snow fronts that will move from the North Sea across Germany in the next few hours.''}
         \label{fig:vis_B}
     \end{subfigure}
    \hfill
     \begin{subfigure}[t]{0.99\textwidth}
         \centering
         \includegraphics[width=\textwidth]{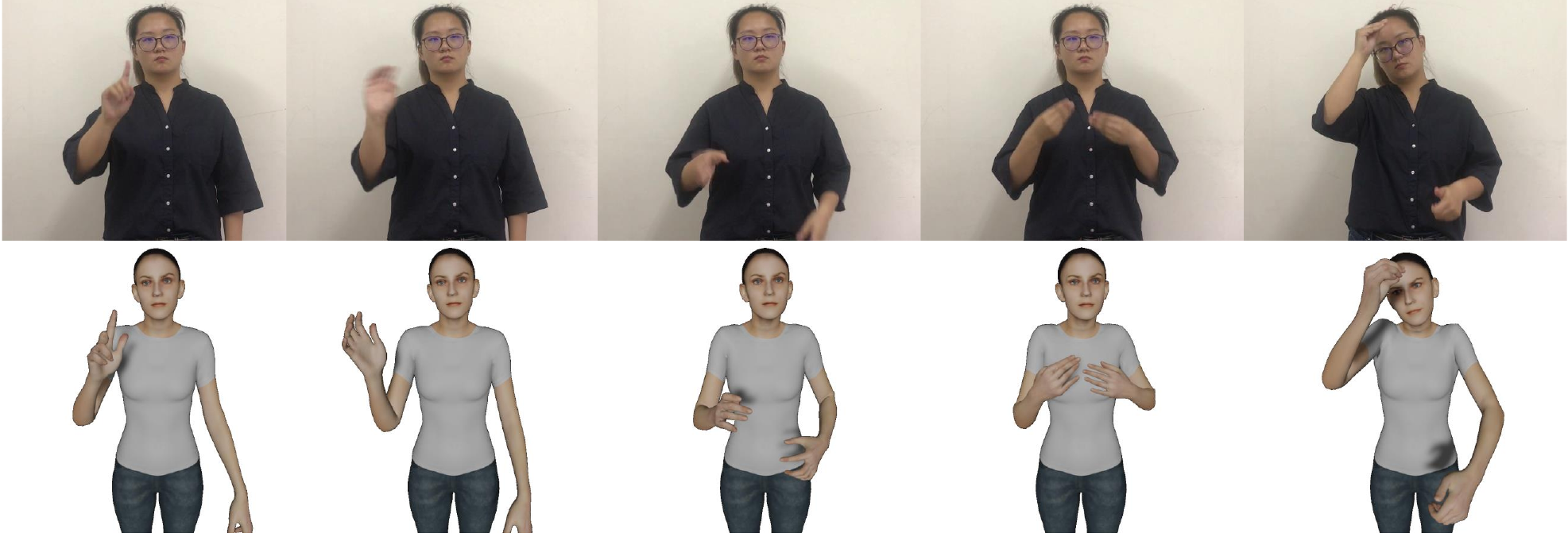}
         \vspace{-4mm}
         \caption{``Do not be a bad example for your brother.''}
         \label{fig:vis_C}
     \end{subfigure}
    
    \vspace{-2mm}
    \caption{Qualitative results on P-2014T~\cite{2014T} (a and b) and CSL~\cite{zhou2021improving} (c). In each sub-figure, we display the text in the caption, and show the ground truth sign video and our translation result in the first row and second row, respectively.}
    \label{fig:vis}
    \vspace{-7mm}
\end{figure}

\vspace{-3mm}
\section{Experiments}
\vspace{-2mm}
\noindent\textbf{Datasets and Evaluation Metrics.} 
Due to the absence of SMPL-X annotations in existing sign language datasets, back-translation is adopted as the primary metric \cite{saunders2022signing, saunders2021mixed, saunders2021continuous} to evaluate the Spoken2Sign system on Phoenix-2014T (P-2014T)~\cite{2014T} and CSL-Daily (CSL)~\cite{zhou2021improving}. We utilize SingleStream-Keypoint \cite{chentwo} as the back translator by default due to its superior keypoint-based Sign2Spoken performance and the availability of its code. P-2014T is a German sign language dataset with a vocabulary size of 1,066 for glosses and 2,887 for German text and there are 7,096/519/642 samples in its training/dev/test set. CSL is a large-scale Chinese sign language dataset, consisting of 18,401/1,077/1,176 samples in the training/dev/test set. Its vocabulary includes 2,000 glosses and 2,343 Chinese words. Following \cite{saunders2022signing, saunders2021mixed, saunders2021continuous}, we report BLEU-4 and ROUGE-L scores. 
In the P-2014T and CSL datasets, signers are recorded while standing stationary in front of the camera, maintaining an almost fixed pose. To verify the effectiveness of 3D keypoint augmentation, we conduct classification experiments on two extra datasets, WLASL \cite{li2020word} and MSASL \cite{joze2019ms}, which feature more pronounced variations in signer poses. We report per-instance/class top-1/5 accuracy~\cite{zuo2023natural}.

\noindent\textbf{Implementation Details.}  Following SMPLify-X~\cite{pavlakos2019expressive}, we optimize Eq.~\ref{eq:smplify-x_improved} for multiple stages with the Limited-memory BFGS optimizer (L-BFGS)~\cite{nocedal2006nonlinear}. By default, we set $\lambda_{1}=3e5$, $\lambda_{2}=7e5$, and $\lambda_{3}=1e3$, in Eq.~\ref{eq:smplify-x_improved}. Each frame takes 300 epochs to fit the SMPL-X model~\cite{pavlakos2019expressive}. To better keep temporal consistency, we pre-estimate the unified shape parameters $\beta$ and global orientation $\zeta$ of SMPL-X and extrinsic camera parameters for the input sign video. When training the sign connector, we set the learning rate as $1e-5$ and use an Adam optimizer \cite{adam}. We filter out extreme samples with too long co-articulations for stable training. 
For a fair comparison with existing works \cite{saunders2022signing, saunders2021mixed, saunders2021continuous, saunders2020progressive, saundersadversarial}, the inputs of all models are the re-projected 2D keypoints.

\vspace{-1mm}
\subsection{Qualitative Evaluation}
\vspace{-1mm}
The objective of this work is to translate spoken languages into sign languages. The translation results are displayed through a 3D avatar. To verify the effectiveness of our system, we show several qualitative results on P-2014T and CSL in Figure~\ref{fig:vis}. It can be seen that the translation results are aesthetically comparable to the ground truth sign videos. Refer to the supplementary materials for more qualitative results.

\vspace{-1mm}
\subsection{Quantitative Evaluation}
\vspace{-1mm}
\label{sec:quantitative}

\begin{table}[t]
\centering
\small
\begin{tabular}{lcccc}
\toprule
\multirow{2}{*}{Method} & \multicolumn{2}{c}{Dev} & \multicolumn{2}{c}{Test} \\
& BLEU-4$\uparrow$ & ROUGE$\uparrow$ & BLEU-4$\uparrow$ & ROUGE$\uparrow$ \\
\midrule
Progressive Transformer \cite{saunders2020progressive} & 11.82 & 33.18 & 10.51 & 32.46 \\
Adversarial Training \cite{saundersadversarial} & 12.65 & 33.68 & 10.81 & 32.74 \\
Mixture Density Networks \cite{saunders2021continuous} & 11.54 & 33.40 & 11.68 & 33.19 \\
Mixture of Motion Primitives \cite{saunders2021mixed} & 14.03 & 37.76 & 13.30 & 36.77 \\
FS-Net \cite{saunders2022signing} & 16.92 & 35.74 & 21.10 & 42.57 \\
SignDiff \cite{fang2023signdiff} & 18.26 & 39.62 & 22.15 & 46.82 \\

\midrule
Ours$^*$ & 22.28 & 47.48 & 22.57 & 48.29 \\
Ours & \textbf{24.16} & \textbf{49.12} & \textbf{25.46} & \textbf{49.68} \\
\bottomrule
\end{tabular}
\caption{Comparison of back-translation performance with existing Spoken2Sign translation works on the P-2014T benchmark. $^*$: using SignDiff's \cite{fang2023signdiff} back translator.}
\vspace{-8mm}
\label{tab:sota}
\end{table}

We employ the widely adopted back-translation metric~\cite{saunders2022signing} to assess the effectiveness of our 3D sign estimator. 
The estimated 3D signs consist of meshes and 3D keypoints. For a fair comparison with previous works, we project these 3D keypoints back into 2D space, adopting a frontal view.
Then, the back translator \cite{chentwo} is trained using these re-projected 2D keypoint sequences to translate sign languages into spoken languages.

\begin{table}[t]
\centering
\resizebox{\linewidth}{!}{
\small
\begin{tabular}{llcccccccc}
\toprule
\multirow{2}{*}{Dataset} & \multirow{2}{*}{Method} & \multicolumn{4}{c}{Dev} & \multicolumn{4}{c}{Test} \\
\cmidrule(r){3-6} \cmidrule(r){7-10}
    & & BLEU-4$\uparrow$ & ROUGE$\uparrow$ & 2D KL$\downarrow$ & TC$\uparrow$ & BLEU-4$\uparrow$ & ROUGE$\uparrow$ & 2D KL$\downarrow$ & TC$\uparrow$ \\
\midrule

\multirow{5}{*}{P-2014T} & \textcolor{gray}{HRNet \cite{wang2020deep}} & \textcolor{gray}{22.94} & \textcolor{gray}{48.81} & \textcolor{gray}{0.00} & \textcolor{gray}{0.961} & \textcolor{gray}{24.95} & \textcolor{gray}{49.13} & \textcolor{gray}{0.00} & \textcolor{gray}{0.962}\\
& SMPLify-X \cite{pavlakos2019expressive} & 19.21 & 44.28 & 31.56 & 0.945 & 19.69 & 43.65 & 31.06 & 0.943 \\
& SMPLer-X$^*$ \cite{cai2023smpler} & 19.49 & 44.95 & 29.50 & 0.966 & 20.01 & 44.93 & 29.72 & 0.965 \\
& OSX$^*$ \cite{lin2023osx} & 22.31 & 47.71 & 26.87 & 0.969 & 23.00 & 47.23 & 27.91 & 0.966 \\
& Ours & \textbf{24.16} & \textbf{49.12} & \textbf{22.09} & \textbf{0.982} & \textbf{25.46} & \textbf{49.68} & \textbf{23.48} & \textbf{0.981} \\
\midrule

\multirow{5}{*}{CSL} & \textcolor{gray}{HRNet \cite{wang2020deep}} & \textcolor{gray}{22.14} & \textcolor{gray}{51.02} & \textcolor{gray}{0.00} & \textcolor{gray}{0.969} & \textcolor{gray}{21.29} & \textcolor{gray}{50.97} & \textcolor{gray}{0.00} & \textcolor{gray}{0.970} \\
& SMPLify-X \cite{pavlakos2019expressive} & 19.31 & 46.46 & 30.69 & 0.947 & 18.96 & 46.71 & 29.66 & 0.949 \\
& SMPLer-X$^*$ \cite{cai2023smpler} & 20.76 & 48.93 & 29.59 & 0.967 & 20.86 & 49.29 & 28.05 & 0.968 \\
& OSX$^*$ \cite{lin2023osx} & 20.44 & 49.00 & 29.75 & 0.965 & 20.29 & 49.60 & 27.40 & 0.965 \\
& Ours & \textbf{21.66} & \textbf{49.69} & \textbf{26.51} & \textbf{0.980} & \textbf{21.44} & \textbf{49.80} & \textbf{24.40} & \textbf{0.981} \\

\bottomrule
\end{tabular}
}
\caption{Comparison with other state-of-the-art 3D sign estimation methods using back-translation. We re-implement SMPLify-X~\cite{pavlakos2019expressive}, SMPLer-X~\cite{cai2023smpler} and OSX~\cite{lin2023osx}. ``HRNet'' denotes a strong baseline that directly estimates 2D keypoints (pseudo ground truth) from raw videos. *: regression-based methods trained on large-scale 3D datasets with SMPL-X annotations. 2D KL: Euclidean distance between re-projected 2D keypoints and pseudo ground truth. TC: temporal consistency measured by calculating the average cosine similarity across all consecutive rendered frames in pixel space.}
\vspace{-9mm}
\label{tab:sign_est}
\end{table}

\noindent\textbf{Comparison with State-of-the-Art Works.}
Table \ref{tab:sota} presents a comparative analysis of our approach against other state-of-the-art works, focusing on back-translation. The prevalent approach in existing works~\cite{saunders2020progressive, saunders2021continuous, saunders2021mixed, saunders2022signing, saundersadversarial, fang2023signdiff} uses generated keypoint sequences to represent sign languages. Additionally, some studies~\cite{saunders2022signing, fang2023signdiff} incorporate a GAN \cite{saunders2021anonysign} or a Diffusion model \cite{zhang2023adding} to further generate synthetic 2D videos from these keypoints. In contrast, our novel Spoken2Sign system produces a 3D sign sequences, visualized through an avatar. In line with the common practice, where back-translation is utilized for evaluation, our approach attains a BLEU-4 score of 24.16 on the P-2014T dev set. It is worth noting that we also re-implement the back translator \cite{tarres2023sign} used in the previous best work, SignDiff \cite{fang2023signdiff}, for a fairer comparison. Although this translator is weaker than our default SingleStream-Keypoint, our approach can still outperform SignDiff by more than 4 BLEU-4 scores on the P-2014T dev set.

\noindent\textbf{3D Sign Estimator.}
In Section~\ref{section: sign estimation}, we present SMPLSign-X, an enhancement of the vanilla SMPLify-X that effectively estimates 3D signs from monocular sign videos. To demonstrate the superior performance of our 3D sign estimator, we benchmark it against three state-of-the-art whole-body estimation methods: SMPLify-X \cite{pavlakos2019expressive}, SMPLer-X \cite{cai2023smpler}, and OSX \cite{lin2023osx}, with the back-translation results detailed in Table \ref{tab:sign_est}. Our approach outperforms all three methods, regardless they are optimization-based (SMPLify-X) or regression-based (SMPLer-X and OSX). Furthermore, our method achieves the lowest 2D keypoint loss and the highest temporal consistency, indicating the precision of the estimated signs and the smoothness of the Spoken2Sign translation results. We attribute these improvements to the incorporation of prior knowledge about sign languages.
Additionally, under the same experimental setting, we introduce a strong baseline that generates 2D keypoints for each frame using a pre-trained HRNet~\cite{wang2020deep}, contrasting with our method and other baselines that re-project the estimated 3D keypoints back into 2D space. We observe that our approach outperforms this baseline on P-2014T. This superiority can be attributed to the consideration of temporal consistency.

\noindent\textbf{User Study with Deaf Participants.}
The primary goal of our Spoken2Sign system is to narrow the communication gap between the deaf and the hearing. Consequently, conducting a user study with deaf participants is essential. We invite four Chinese signers to evaluate our Spoken2Sign results on three aspects: naturalness (checking for awkward poses), smoothness (observing for noticeable shaking between frames), and similarity to raw videos\footnote{Though the participants are not native in German sign language, their familiarity with general sign language rules enables them to evaluate the similarity to sign videos for P-2014T.}. We provide 100 randomly selected videos from each dataset to the participants, who then give a rating ranging from 1 to 5 for each aspect. The average ratings across all videos and participants are reported in Table \ref{tab:user_study}. It is evident that our approach significantly outperforms the baseline, SMPLify-X \cite{pavlakos2019expressive}, in all aspects.

\noindent\textbf{Loss Functions.}
In Section \ref{section: sign estimation}, we introduce three loss functions designed to facilitate 3D sign estimation: $\mathcal{L}_{unseen}$, which draws the unseen keypoints closer to those of the rest pose; $\mathcal{L}_{upright}$, aimed at encouraging an upright posture in the upper body; and $\mathcal{L}_{smooth}$, which preserves temporal consistency to generate visually appealing results. As shown in Table \ref{tab:loss_abl}, omitting any of these loss functions leads to degraded performance.

\begin{table}[t]
    \begin{minipage}{.43\linewidth}
    \centering
    \resizebox{1.0\linewidth}{!}{
    \begin{tabular}{llccc}
    \toprule
    Dataset & Method & Nat.$\uparrow$ & Smo.$\uparrow$ & Sim.$\uparrow$ \\
    \midrule
    \multirow{2}{*}{P-2014T} & SMPLify-X \cite{pavlakos2019expressive} & 1.52 & 1.98 & 2.41 \\
    & Ours & \textbf{3.58} & \textbf{4.04} & \textbf{3.94} \\
    \midrule
    
    \multirow{2}{*}{CSL} & SMPLify-X \cite{pavlakos2019expressive} & 1.27 & 1.75 & 1.69 \\
    & Ours & \textbf{3.78} & \textbf{4.14} & \textbf{3.78} \\
    \bottomrule
    \end{tabular}
    }
    \caption{User study with deaf participants. Nat.: naturalness; Smo.: smoothness; Sim.: similarity to the raw video. Score range: 1-5.}
    \label{tab:user_study}
    \end{minipage}
    \hfill
    \begin{minipage}{.54\linewidth}
    \centering
    \resizebox{1.0\linewidth}{!}{
    \begin{tabular}{lcccc}
    \toprule
    \multirow{2}{*}{Method} & \multicolumn{2}{c}{Dev} & \multicolumn{2}{c}{Test} \\
    & BLEU-4$\uparrow$ & ROUGE$\uparrow$ & BLEU-4$\uparrow$ & ROUGE$\uparrow$ \\
    
    \midrule
    w/o $\mathcal{L}_{unseen}$ & 22.57 & 47.25 & 23.70 & 47.62 \\
    w/o $\mathcal{L}_{upright}$ & 23.10 & 47.93 & 23.99 & 48.00 \\
    w/o $\mathcal{L}_{smooth}$ & 23.64 & 48.50 & 24.36 & 48.35 \\
    w/ all (default) & \textbf{24.16} & \textbf{49.12} & \textbf{25.46} & \textbf{49.68} \\
    
    \bottomrule
    \end{tabular}
    }
    \caption{Ablation study for the proposed loss functions, $\mathcal{L}_{unseen}$, $\mathcal{L}_{upright}$, and $\mathcal{L}_{smooth}$, on P-2014T using back-translation.}
    \label{tab:loss_abl}
    \end{minipage}
\vspace{-8mm}
\end{table}

\begin{table}[t]
    \begin{minipage}{.43\linewidth}
    \centering
    \resizebox{1.0\linewidth}{!}{
    \begin{tabular}{lcccc}
    \toprule
    \multirow{2}{*}{Method} & \multirow{2}{*}{\shortstack{Fixed\\Duration}} & \multirow{2}{*}{\shortstack{Hand \\Only}} & \multirow{2}{*}{\shortstack{w/o\\CD}} & \multirow{2}{*}{Default} \\
    & \\
    \midrule
    L1 distance$\downarrow$ & 1.83 & 1.22 & 1.34 & \textbf{1.04} \\
    
    \bottomrule
    \end{tabular}
    }
    \caption{Ablation study for the design of sign connector on the P-2014T dev set. CD: coordinate distance.}
    \label{tab:sign_connector}
    \end{minipage}
    \hfill
    \begin{minipage}{.54\linewidth}
    \centering
    \resizebox{1.0\linewidth}{!}{
    \begin{tabular}{lcccc}
    \toprule
    \multirow{2}{*}{Method} & \multicolumn{2}{c}{Dev} & \multicolumn{2}{c}{Test} \\
    & BLEU-4$\uparrow$ & ROUGE$\uparrow$ & BLEU-4$\uparrow$ & ROUGE$\uparrow$ \\
    \midrule
    2D connector & 21.83 & 46.87 & 22.03 & 47.43 \\
    \midrule
    w/o connector & 20.69 & 46.13 & 20.82 & 44.96 \\
    w/o retrieval & 22.25 & 46.70 & 23.81 & 48.08 \\
    w/ both (default) & \textbf{24.16} & \textbf{49.12} & \textbf{25.46} & \textbf{49.68} \\
    \bottomrule
    \end{tabular}
    }
    \caption{Ablation study for sign connector and retrieval on P-2014T using back-translation.}
    \label{tab:abl_conn_ret}
    \end{minipage}
\vspace{-9mm}
\end{table}

\noindent\textbf{Sign Connector and Retrieval.} 
The objective of our sign connector (see Figure \ref{fig:conn}) is to predict the duration of the co-articulation and stitch two adjacent 3D signs. The default configuration of our sign connector takes a concatenation of the 3D keypoints of the preceding sign $\mathcal{D}^{pre}_{\mathcal{J}_{SC}}$, the 3D keypoints of the succeeding sign $\mathcal{D}^{next}_{\mathcal{J}_{SC}}$, and their coordinate distance $\mathcal{D}^{pre}_{\mathcal{J}_{SC}} - \mathcal{D}^{next}_{\mathcal{J}_{SC}}$, as inputs. In Table~\ref{tab:sign_connector}, we compare this default approach with a baseline, where the duration of each co-articulation is set to a fixed value of 4 (the average duration derived from the training set's statistics). We also consider two variants: ``hand only'' and ``without coordinate distance'' (w/o CD). The ``hand only'' variant uses only hand keypoints as inputs, while the ``w/o CD'' variant relies solely on the concatenation of $\mathcal{D}^{pre}_{\mathcal{J}_{SC}}$ and $\mathcal{D}^{next}_{\mathcal{J}_{SC}}$. For each strategy, we calculate the average L1 distance between the predictions and ground truths across all co-articulations. Our default strategy outperforms all other variants.

We also delve into the effects of sign connector and sign retrieval on the back-translation efficacy of the entire system. First, we modify our sign connector to stitch signs within the 2D space, as opposed to the default 3D space. As indicated in Table \ref{tab:abl_conn_ret}, the observed decrease in performance highlights the importance of modeling co-articulations in 3D space. Furthermore, removing either the connector (\ie, directly concatenating signs) or the retriever (\ie, randomly selecting a sign for each gloss) also leads to worse back-translation results.

\begin{table}[t]
    \begin{minipage}{.43\linewidth}
    \centering
    \resizebox{1.0\linewidth}{!}{
    \begin{tabular}{lccccc}
    \toprule
    \multirow{2}{*}{Dataset} & 3D Kp. & \multicolumn{2}{c}{Per-instance} & \multicolumn{2}{c}{Per-class} \\
    & Aug. & Top-1$\uparrow$ & Top-5$\uparrow$ & Top-1$\uparrow$ & Top-5$\uparrow$ \\
    \midrule
    
    \multirow{2}{*}{WLASL} &  & 46.20 & 78.81 & 43.72 & 77.55 \\
    & \checkmark & \textbf{47.66} & \textbf{79.71} & \textbf{45.10} & \textbf{78.16} \\
    \midrule
    
    \multirow{2}{*}{MSASL} &  & 51.81 & 73.78 & 48.52 & 71.76 \\
    & \checkmark & \textbf{53.10} & \textbf{75.06} & \textbf{49.71} & \textbf{72.71} \\
    
    \bottomrule
    \end{tabular}
    }
    \caption{Ablation study on 3D keypoint (Kp.) augmentation (Aug.).}
    \label{tab:pose_aug}
    \end{minipage}
    \hfill
    \begin{minipage}{.55\linewidth}
    \centering
    \resizebox{1.0\linewidth}{!}{
    \begin{tabular}{lcccccc}
    \toprule
    \multirow{2}{*}{Dataset} & \multicolumn{2}{c}{View} & \multicolumn{2}{c}{Dev} & \multicolumn{2}{c}{Test} \\
    & Frontal & Side & BLEU-4$\uparrow$ & ROUGE$\uparrow$ & BLEU-4$\uparrow$ & ROUGE$\uparrow$ \\
    \midrule
    
    \multirow{3}{*}{P-2014T} & \checkmark & & 24.16 & 49.12 & 25.46 & 49.68 \\
    & & \checkmark & 22.60 & 47.31 & 23.49 & 47.31 \\
    & \checkmark & \checkmark & \textbf{24.69} & \textbf{50.34} & \textbf{26.54} & \textbf{50.69} \\
    \midrule
    
    \multirow{3}{*}{CSL} & \checkmark & & 21.66 & 49.69 & 21.44 & 49.80 \\
    & & \checkmark & 20.44 & 48.31 & 20.03 & 48.60 \\
    & \checkmark & \checkmark & \textbf{23.14} & \textbf{52.13} & \textbf{22.22} & \textbf{51.47} \\
    
    \bottomrule
    \end{tabular}
    }
    \caption{Ablation study on multi-view Sign2Spoken translation.}
    \label{tab:side_view}
    \end{minipage}
\vspace{-9mm}
\end{table}

\vspace{-3mm}
\subsection{Effectiveness of the By-Products}
\vspace{-1mm}
Our approach is able to estimate 3D signs from monocular sign videos.
In Section~\ref{sec:app}, we introduce two by-products, 3D keypoint augmentation and multi-view understanding, which have the potential to improve keypoint-based models for sign language understanding.

\noindent\textbf{3D Keypoint Augmentation.}
3D signs can be rotated at any angle, motivating us to develop 3D keypoint augmentation. This is achieved by randomly rotating the input 3D sign by a small angle before projecting it into 2D space. To validate its effectiveness, we turn to two challenging ISLR datasets, WLASL and MSASL, where the variation of signer poses is more dramatic than that in P-2014T and CSL. We use NLA-SLR-Keypoint-64~\cite{zuo2023natural} as the base model. As shown in Table \ref{tab:pose_aug}, 3D keypoint augmentation consistently improves model performance across all metrics on the dev sets of both datasets, with almost no extra training cost.

\noindent\textbf{Multi-View Spoken2Sign Translation.}
Another by-product is the use of side-view keypoints to enhance the performance of the model trained on frontal-view keypoints. Table \ref{tab:side_view} shows the effectiveness of multi-view Spoken2Sign translation, with the SingleStream-Keypoint~\cite{chentwo} as the base model.

\vspace{-2mm}
\section{Conclusion}
\vspace{-2mm}
This paper focuses on Spoken2Sign translation, a reverse process to traditional Sign2Spoken translation, aimed at narrowing the communication gap between deaf and hearing individuals. In contrast to prior works that produce translation results in 2D space, our innovative method generates 3D signs using the proposed techniques such as SMPLSign-X and sign connector. The translation results are displayed through an avatar. Our method involves three main steps: 1) building a sign language dictionary; 2) estimating the 3D representation for each sign in this dictionary; and 3) executing Spoken2Sign translation and rendering an avatar. Additionally, we introduce two by-products, 3D keypoint augmentation and multi-view understanding, which significantly promote the keypoint-based models. Extensive experiments demonstrate the effectiveness of our approach.

%
%
\bibliographystyle{splncs04}
\bibliography{main}

\begin{thebibliography}{10}
\providecommand{\url}[1]{\texttt{#1}}
\providecommand{\urlprefix}{URL }
\providecommand{\doi}[1]{https://doi.org/#1}

\bibitem{albanie2020bsl}
Albanie, S., Varol, G., Momeni, L., Afouras, T., Chung, J.S., Fox, N., Zisserman, A.: {BSL-1K}: Scaling up co-articulated sign language recognition using mouthing cues. In: ECCV. pp. 35--53 (2020)

\bibitem{bao2017cvae}
Bao, J., Chen, D., Wen, F., Li, H., Hua, G.: Cvae-gan: fine-grained image generation through asymmetric training. In: ICCV. pp. 2745--2754 (2017)

\bibitem{cai2023smpler}
Cai, Z., Yin, W., Zeng, A., Wei, C., Sun, Q., Wang, Y., Pang, H.E., Mei, H., Zhang, M., Zhang, L., et~al.: Smpler-x: Scaling up expressive human pose and shape estimation. In: NeurIPS (2023)

\bibitem{2014T}
Camgoz, N.C., Hadfield, S., Koller, O., Ney, H., Bowden, R.: Neural sign language translation. In: CVPR. pp. 7784--7793 (2018)

\bibitem{camgoz2020multi}
Camgoz, N.C., Koller, O., Hadfield, S., Bowden, R.: Multi-channel transformers for multi-articulatory sign language translation. In: ECCV. pp. 301--319. Springer (2020)

\bibitem{slt}
Camg{\"{o}}z, N.C., Koller, O., Hadfield, S., Bowden, R.: Sign language transformers: Joint end-to-end sign language recognition and translation. In: CVPR. pp. 10020--10030 (2020)

\bibitem{chen2020simple}
Chen, T., Kornblith, S., Norouzi, M., Hinton, G.: A simple framework for contrastive learning of visual representations. In: ICML. pp. 1597--1607. PMLR (2020)

\bibitem{chen2022simple}
Chen, Y., Wei, F., Sun, X., Wu, Z., Lin, S.: A simple multi-modality transfer learning baseline for sign language translation. In: CVPR. pp. 5120--5130 (2022)

\bibitem{chentwo}
Chen, Y., Zuo, R., Wei, F., Wu, Y., Liu, S., Mak, B.: Two-stream network for sign language recognition and translation. In: NeurIPS (2022)

\bibitem{fcn}
Cheng, K.L., Yang, Z., Chen, Q., Tai, Y.: Fully convolutional networks for continuous sign language recognition. In: ECCV. vol. 12369, pp. 697--714 (2020)

\bibitem{cheng2023cico}
Cheng, Y., Wei, F., Bao, J., Chen, D., Zhang, W.: Cico: Domain-aware sign language retrieval via cross-lingual contrastive learning. In: CVPR (2023)

\bibitem{blender}
Community, B.O.: Blender - a 3D modelling and rendering package. Blender Foundation, Stichting Blender Foundation, Amsterdam (2018), \url{http://www.blender.org}

\bibitem{dnf}
Cui, R., Liu, H., Zhang, C.: A deep neural framework for continuous sign language recognition by iterative training. IEEE TMM  \textbf{PP}, ~1--1 (07 2019)

\bibitem{duarte2022sign}
Duarte, A., Albanie, S., Gir{\'o}-i Nieto, X., Varol, G.: Sign language video retrieval with free-form textual queries. In: CVPR. pp. 14094--14104 (2022)

\bibitem{duarte2021how2sign}
Duarte, A., Palaskar, S., Ventura, L., Ghadiyaram, D., DeHaan, K., Metze, F., Torres, J., Giro-i Nieto, X.: How2sign: a large-scale multimodal dataset for continuous american sign language. In: CVPR. pp. 2735--2744 (2021)

\bibitem{fang2023signdiff}
Fang, S., Sui, C., Zhang, X., Tian, Y.: Signdiff: Learning diffusion models for american sign language production (2023)

\bibitem{forte2023reconstructing}
Forte, M.P., Kulits, P., Huang, C.H.P., Choutas, V., Tzionas, D., Kuchenbecker, K.J., Black, M.J.: Reconstructing signing avatars from video using linguistic priors. In: CVPR. pp. 12791--12801 (2023)

\bibitem{gan2023contrastive}
Gan, S., Yin, Y., Jiang, Z., Xia, K., Xie, L., Lu, S.: Contrastive learning for sign language recognition and translation. In: IJCAI. pp. 763--772 (2023)

\bibitem{geman1987statistical}
Geman, S.: Statistical methods for tomographic image reconstruction. Bulletin of International Statistical Institute  \textbf{4},  5--21 (1987)

\bibitem{ctc}
Graves, A., Fern{\'a}ndez, S., Gomez, F., Schmidhuber, J.: Connectionist temporal classification: labelling unsegmented sequence data with recurrent neural networks. In: ICML. pp. 369--376 (2006)

\bibitem{self-mutual}
Hao, A., Min, Y., Chen, X.: Self-mutual distillation learning for continuous sign language recognition. In: ICCV. pp. 11303--11312 (October 2021)

\bibitem{hu2023signbert+}
Hu, H., Zhao, W., Zhou, W., Li, H.: Signbert+: Hand-model-aware self-supervised pre-training for sign language understanding. IEEE TPAMI  (2023)

\bibitem{hu2021hand}
Hu, H., Zhou, W., Li, H.: Hand-model-aware sign language recognition. In: AAAI. vol.~35, pp. 1558--1566 (2021)

\bibitem{hu2022self}
Hu, L., Gao, L., Feng, W., et~al.: Self-emphasizing network for continuous sign language recognition. In: AAAI (2023)

\bibitem{hu2023continuous}
Hu, L., Gao, L., Liu, Z., Feng, W.: Continuous sign language recognition with correlation network. In: CVPR (2023)

\bibitem{huang2021towards}
Huang, W., Pan, W., Zhao, Z., Tian, Q.: Towards fast and high-quality sign language production. In: MM. pp. 3172--3181 (2021)

\bibitem{huang2022dualsign}
Huang, W., Zhao, Z., He, J., Zhang, M.: Dualsign: Semi-supervised sign language production with balanced multi-modal multi-task dual transformation. In: MM. pp. 5486--5495 (2022)

\bibitem{hwang2021non}
Hwang, E., Kim, J.H., Park, J.C.: Non-autoregressive sign language production with gaussian space. In: BMVC (2021)

\bibitem{jiang2021skeleton}
Jiang, S., Sun, B., Wang, L., Bai, Y., Li, K., Fu, Y.: Skeleton aware multi-modal sign language recognition. In: CVPRW. pp. 3413--3423 (2021)

\bibitem{jin2020whole}
Jin, S., Xu, L., Xu, J., Wang, C., Liu, W., Qian, C., Ouyang, W., Luo, P.: Whole-body human pose estimation in the wild. In: ECCV. pp. 196--214 (2020)

\bibitem{joo2018total}
Joo, H., Simon, T., Sheikh, Y.: Total capture: A 3d deformation model for tracking faces, hands, and bodies. In: CVPR. pp. 8320--8329 (2018)

\bibitem{joze2019ms}
Joze, H.R.V., Koller, O.: {MS-ASL}: A large-scale data set and benchmark for understanding {American} sign language. In: BMVC (2019)

\bibitem{kanazawa2018end}
Kanazawa, A., Black, M.J., Jacobs, D.W., Malik, J.: End-to-end recovery of human shape and pose. In: CVPR. pp. 7122--7131 (2018)

\bibitem{adam}
Kingma, D.P., Ba, J.: Adam: {A} method for stochastic optimization. In: ICLR (2015)

\bibitem{slp_avatar_2}
Kipp, M., Heloir, A., Nguyen, Q.: Sign language avatars: Animation and comprehensibility. In: Intelligent Virtual Agents: 10th International Conference, IVA 2011, Reykjavik, Iceland, September 15-17, 2011. Proceedings 11. pp. 113--126. Springer (2011)

\bibitem{kudo2018sentencepiece}
Kudo, T., Richardson, J.: Sentencepiece: A simple and language independent subword tokenizer and detokenizer for neural text processing. arXiv preprint arXiv:1808.06226  (2018)

\bibitem{kurzinger2020ctc}
K{\"u}rzinger, L., Winkelbauer, D., Li, L., Watzel, T., Rigoll, G.: Ctc-segmentation of large corpora for german end-to-end speech recognition. In: International Conference on Speech and Computer. pp. 267--278. Springer (2020)

\bibitem{lassner2017unite}
Lassner, C., Romero, J., Kiefel, M., Bogo, F., Black, M.J., Gehler, P.V.: Unite the people: Closing the loop between 3d and 2d human representations. In: CVPR. pp. 6050--6059 (2017)

\bibitem{lee2023human}
Lee, T., Oh, Y., Lee, K.M.: Human part-wise 3d motion context learning for sign language recognition. In: ICCV. pp. 20740--20750 (2023)

\bibitem{li2020word}
Li, D., Rodriguez, C., Yu, X., Li, H.: Word-level deep sign language recognition from video: A new large-scale dataset and methods comparison. In: WACV. pp. 1459--1469 (2020)

\bibitem{li2020tspnet}
Li, D., Xu, C., Yu, X., Zhang, K., Swift, B., Suominen, H., Li, H.: Tspnet: Hierarchical feature learning via temporal semantic pyramid for sign language translation. In: NeurIPS. vol.~33, pp. 12034--12045 (2020)

\bibitem{li2020transferring}
Li, D., Yu, X., Xu, C., Petersson, L., Li, H.: Transferring cross-domain knowledge for video sign language recognition. In: CVPR. pp. 6205--6214 (2020)

\bibitem{lin2023osx}
Lin, J., Zeng, A., Wang, H., Zhang, L., Li, Y.: One-stage 3d whole-body mesh recovery with component aware transformer. In: CVPR. pp. 21159--21168 (2023)

\bibitem{liu2020multilingual}
Liu, Y., Gu, J., Goyal, N., Li, X., Edunov, S., Ghazvininejad, M., Lewis, M., Zettlemoyer, L.: Multilingual denoising pre-training for neural machine translation. TACL  \textbf{8},  726--742 (2020)

\bibitem{slp_avatar_1}
McDonald, J., Wolfe, R., Schnepp, J., Hochgesang, J., Jamrozik, D.G., Stumbo, M., Berke, L., Bialek, M., Thomas, F.: An automated technique for real-time production of lifelike animations of american sign language. Universal Access in the Information Society  \textbf{15},  551--566 (2016)

\bibitem{vac}
Min, Y., Hao, A., Chai, X., Chen, X.: Visual alignment constraint for continuous sign language recognition. In: ICCV. pp. 11542--11551 (October 2021)

\bibitem{momeni2022automatic}
Momeni, L., Bull, H., Prajwal, K., Albanie, S., Varol, G., Zisserman, A.: Automatic dense annotation of large-vocabulary sign language videos. In: ECCV. pp. 671--690 (2022)

\bibitem{momeni2020watch}
Momeni, L., Varol, G., Albanie, S., Afouras, T., Zisserman, A.: Watch, read and lookup: learning to spot signs from multiple supervisors. In: ACCV (2020)

\bibitem{niu2024hong}
Niu, Z., Zuo, R., Mak, B., Wei, F.: A hong kong sign language corpus collected from sign-interpreted tv news. In: LREC-COLING. pp. 636--646 (2024)

\bibitem{nocedal2006nonlinear}
Nocedal, J., Wright, S.J.: Nonlinear equations. Numerical Optimization pp. 270--302 (2006)

\bibitem{pavlakos2019expressive}
Pavlakos, G., Choutas, V., Ghorbani, N., Bolkart, T., Osman, A.A., Tzionas, D., Black, M.J.: Expressive body capture: 3d hands, face, and body from a single image. In: CVPR. pp. 10975--10985 (2019)

\bibitem{pratap2023scaling}
Pratap, V., Tjandra, A., Shi, B., Tomasello, P., Babu, A., Kundu, S., Elkahky, A., Ni, Z., Vyas, A., Fazel-Zarandi, M., et~al.: Scaling speech technology to 1,000+ languages. arXiv preprint arXiv:2305.13516  (2023)

\bibitem{slp_review}
Rastgoo, R., Kiani, K., Escalera, S., Sabokrou, M.: Sign language production: A review. In: CVPRW. pp. 3451--3461 (2021)

\bibitem{razavi2019generating}
Razavi, A., Van~den Oord, A., Vinyals, O.: Generating diverse high-fidelity images with vq-vae-2. NeurIPS  \textbf{32} (2019)

\bibitem{sak2015learning}
Sak, H., Senior, A., Rao, K., Irsoy, O., Graves, A., Beaufays, F., Schalkwyk, J.: Learning acoustic frame labeling for speech recognition with recurrent neural networks. In: ICASSP. pp. 4280--4284 (2015)

\bibitem{saundersadversarial}
Saunders, B., Camg{\"o}z, N.C., Bowden, R.: Adversarial training for multi-channel sign language production. In: BMVC (2020)

\bibitem{saunders2020progressive}
Saunders, B., Camgoz, N.C., Bowden, R.: Progressive transformers for end-to-end sign language production. In: ECCV. pp. 687--705 (2020)

\bibitem{saunders2021anonysign}
Saunders, B., Camgoz, N.C., Bowden, R.: Anonysign: Novel human appearance synthesis for sign language video anonymisation. In: FG 2021. pp.~1--8 (2021)

\bibitem{saunders2021continuous}
Saunders, B., Camgoz, N.C., Bowden, R.: Continuous 3d multi-channel sign language production via progressive transformers and mixture density networks. IJCV  \textbf{129}(7),  2113--2135 (2021)

\bibitem{saunders2021mixed}
Saunders, B., Camgoz, N.C., Bowden, R.: Mixed signals: Sign language production via a mixture of motion primitives. In: ICCV. pp. 1919--1929 (2021)

\bibitem{saunders2022signing}
Saunders, B., Camgoz, N.C., Bowden, R.: Signing at scale: Learning to co-articulate signs for large-scale photo-realistic sign language production. In: CVPR. pp. 5141--5151 (2022)

\bibitem{shmelkov2018good}
Shmelkov, K., Schmid, C., Alahari, K.: How good is my gan? In: ECCV. pp. 213--229 (2018)

\bibitem{stoll2018sign}
Stoll, S., Camg{\"o}z, N.C., Hadfield, S., Bowden, R.: Sign language production using neural machine translation and generative adversarial networks. In: BMVC (2018)

\bibitem{stoll2020text2sign}
Stoll, S., Camgoz, N.C., Hadfield, S., Bowden, R.: Text2sign: towards sign language production using neural machine translation and generative adversarial networks. IJCV  \textbf{128}(4),  891--908 (2020)

\bibitem{tang2022gloss}
Tang, S., Hong, R., Guo, D., Wang, M.: Gloss semantic-enhanced network with online back-translation for sign language production. In: MM. pp. 5630--5638 (2022)

\bibitem{tarres2023sign}
Tarr{\'e}s, L., G{\'a}llego, G.I., Duarte, A., Torres, J., Gir{\'o}-i Nieto, X.: Sign language translation from instructional videos. In: CVPRW. pp. 5624--5634 (2023)

\bibitem{varol2021read}
Varol, G., Momeni, L., Albanie, S., Afouras, T., Zisserman, A.: Read and attend: Temporal localisation in sign language videos. In: CVPR. pp. 16857--16866 (2021)

\bibitem{wang2020deep}
Wang, J., Sun, K., Cheng, T., Jiang, B., Deng, C., Zhao, Y., Liu, D., Mu, Y., Tan, M., Wang, X., et~al.: Deep high-resolution representation learning for visual recognition. IEEE TPAMI  \textbf{43}(10),  3349--3364 (2020)

\bibitem{Wei_2023_ICCV}
Wei, F., Chen, Y.: Improving continuous sign language recognition with cross-lingual signs. In: ICCV. pp. 23612--23621 (October 2023)

\bibitem{xiang2019monocular}
Xiang, D., Joo, H., Sheikh, Y.: Monocular total capture: Posing face, body, and hands in the wild. In: CVPR. pp. 10965--10974 (2019)

\bibitem{xie2023vector}
Xie, P., Zhang, Q., Li, Z., Tang, H., Du, Y., Hu, X.: Vector quantized diffusion model with codeunet for text-to-sign pose sequences generation (2023)

\bibitem{xu2020ghum}
Xu, H., Bazavan, E.G., Zanfir, A., Freeman, W.T., Sukthankar, R., Sminchisescu, C.: Ghum \& ghuml: Generative 3d human shape and articulated pose models. In: CVPR. pp. 6184--6193 (2020)

\bibitem{xu2019denserac}
Xu, Y., Zhu, S.C., Tung, T.: Denserac: Joint 3d pose and shape estimation by dense render-and-compare. In: ICCV. pp. 7760--7770 (2019)

\bibitem{Yao_2023_ICCV}
Yao, H., Zhou, W., Feng, H., Hu, H., Zhou, H., Li, H.: Sign language translation with iterative prototype. In: ICCV. pp. 15592--15601 (October 2023)

\bibitem{yin2021simulslt}
Yin, A., Zhao, Z., Liu, J., Jin, W., Zhang, M., Zeng, X., He, X.: Simulslt: End-to-end simultaneous sign language translation. In: MM. pp. 4118--4127 (2021)

\bibitem{yin2020better}
Yin, K., Read, J.: Better sign language translation with stmc-transformer. COLING  (2020)

\bibitem{yu2023efficient}
Yu, P., Zhang, L., Fu, B., Chen, Y.: Efficient sign language translation with a curriculum-based non-autoregressive decoder. In: IJCAI. pp. 5260--5268 (2023)

\bibitem{zelinka2020neural}
Zelinka, J., Kanis, J.: Neural sign language synthesis: Words are our glosses. In: WACV. pp. 3395--3403 (2020)

\bibitem{zhang2023adding}
Zhang, L., Rao, A., Agrawala, M.: Adding conditional control to text-to-image diffusion models. In: ICCV. pp. 3836--3847 (2023)

\bibitem{best}
Zhao, W., Hu, H., Zhou, W., Shi, J., Li, H.: {BEST:} {BERT} pre-training for sign language recognition with coupling tokenization. In: AAAI (2023)

\bibitem{zheng2023cvt}
Zheng, J., Wang, Y., Tan, C., Li, S., Wang, G., Xia, J., Chen, Y., Li, S.Z.: {CVT-SLR}: Contrastive visual-textual transformation for sign language recognition with variational alignment. In: CVPR (2023)

\bibitem{Zhou_2023_ICCV}
Zhou, B., Chen, Z., Clap\'es, A., Wan, J., Liang, Y., Escalera, S., Lei, Z., Zhang, D.: Gloss-free sign language translation: Improving from visual-language pretraining. In: ICCV. pp. 20871--20881 (October 2023)

\bibitem{zhou2021improving}
Zhou, H., Zhou, W., Qi, W., Pu, J., Li, H.: Improving sign language translation with monolingual data by sign back-translation. In: CVPR. pp. 1316--1325 (2021)

\bibitem{stmc}
Zhou, H., Zhou, W., Zhou, Y., Li, H.: Spatial-temporal multi-cue network for continuous sign language recognition. In: AAAI. pp. 13009--13016 (2020)

\bibitem{zuo2022c2slr}
Zuo, R., Mak, B.: {C2SLR}: Consistency-enhanced continuous sign language recognition. In: CVPR. pp. 5131--5140 (2022)

\bibitem{zuo2024improving}
Zuo, R., Mak, B.: Improving continuous sign language recognition with consistency constraints and signer removal. ACM TOMM  \textbf{20}(6),  1--25 (2024)

\bibitem{zuo2023natural}
Zuo, R., Wei, F., Mak, B.: Natural language-assisted sign language recognition. In: CVPR (2023)

\bibitem{zuo2024towards}
Zuo, R., Wei, F., Mak, B.: Towards online sign language recognition and translation. arXiv preprint arXiv:2401.05336  (2024)

\end{thebibliography}

\appendix
\section{More Implementation Details}

\noindent\textbf{Dictionary Construction.} 
As described in Section 3.1 of the main paper, a well-optimized CSLR model, TwoStream-SLR \cite{chentwo}, serves as our sign segmentor to segment a given continuous sign into several isolated ones. Below, we formulate the entire process.

Given a continuous sign video $\boldsymbol{V}$ with $T$ frames, and its corresponding ground truth gloss sequence $\boldsymbol{g}=(g_1,\dots, g_{N})$ containing $N$ glosses, the probability of an alignment path $\boldsymbol{\theta} = (\theta_1, \dots, \theta_T)$ with respect to the ground truth $\boldsymbol{g}$, where $\theta_t \in \{g_i\}_{i=1}^N \cup \{background\}$, can be calculated by:
\begin{equation}
\label{eq:cond_prob}
    p(\boldsymbol{\theta}|\boldsymbol{V}) = \prod_{t=1}^T p_t(\theta_t),
\end{equation}
where $p_t(\theta_t)$ denotes the posterior probability of predicting the $t$-th frame as class $\theta_t$. The optimal path $\boldsymbol{\theta}^*$ is the one with the maximum probability in the set of all feasible alignment paths $\mathcal{S}(\boldsymbol{g})$ with respect to the ground truth $\boldsymbol{g}$: 
\begin{equation}
    \boldsymbol{\theta}^* = \argmax_{\boldsymbol{\theta} \in \mathcal{S}(\boldsymbol{g})} p(\boldsymbol{\theta}|\boldsymbol{V}).
\end{equation}
The optimal path $\boldsymbol{\theta}^*$ can be efficiently identified using the CTC forced alignment algorithm \cite{dnf,ctc,Wei_2023_ICCV}, a well-established technique in the speech community to accurately align transcripts to speech signals \cite{kurzinger2020ctc,pratap2023scaling,sak2015learning}. Subsequently, we aggregate successive duplicate predictions into a single isolated sign.

\noindent\textbf{Text2Gloss Translator.} 
We utilize mBART~\cite{liu2020multilingual}, a pre-trained sequence-to-sequence denoising auto-encoder, as our Text2Gloss translator. This model adopts a standard Transformer architecture, featuring 12 encoder and decoder layers, which ensures a robust contextual understanding. Our mBART model is initialized with the one pre-trained on a large multilingual corpus. To tailor mBART for Text2Gloss translation, we tokenize the gloss sequences and text sentences into sub-word units using the SentencePiece tokenizer~\cite{kudo2018sentencepiece} and incorporate positional embeddings. We train the model for 80 epochs, starting with an initial learning rate of $1e-5$, and apply a dropout rate of 0.3 and label smoothing with a factor of 0.2 to prevent overfitting.

\section{More Qualitative Results}
\begin{figure}[!ht]
     \centering
     \begin{subfigure}[t]{0.99\textwidth}
         \centering
         \includegraphics[width=\textwidth]{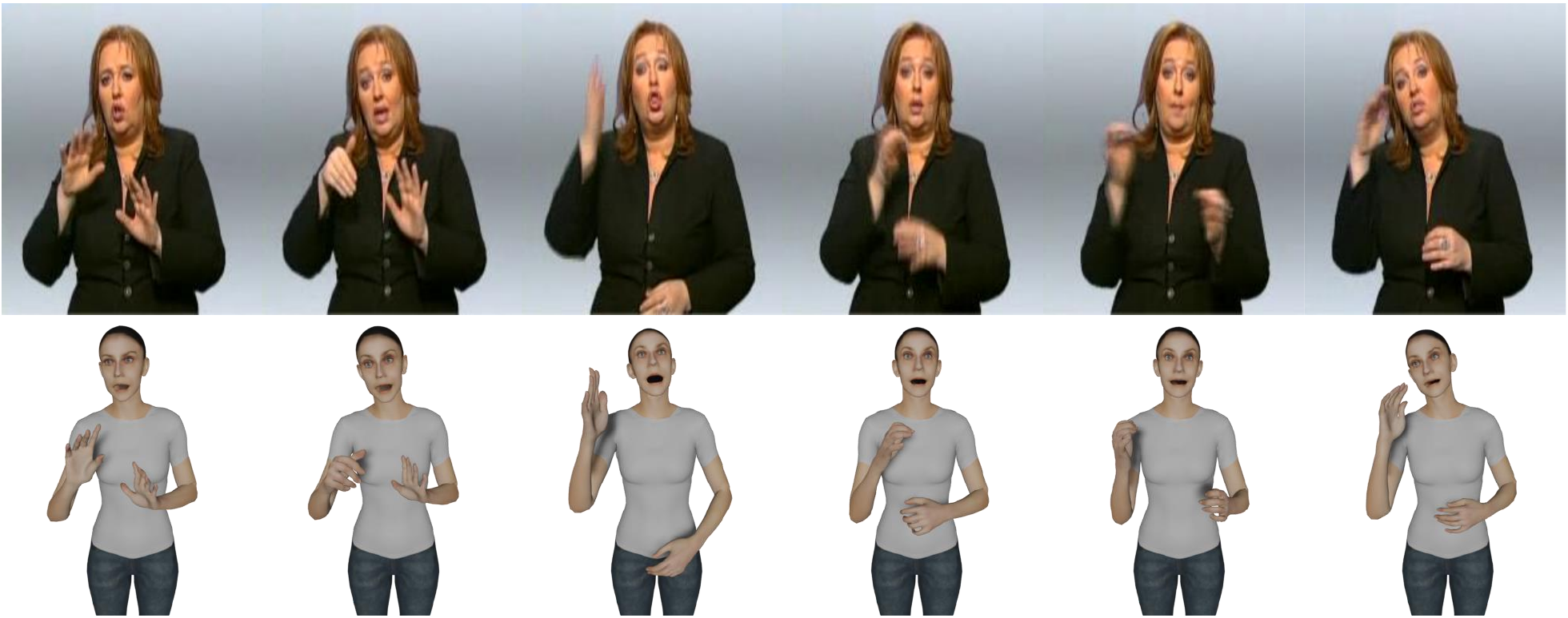}
         \caption{``In the south there is a weak wind. In the north it blows moderately on the coasts with strong to stormy gusts.''}
         \label{fig:supp_phoenix_A}
     \end{subfigure}
     \hfill
    \begin{subfigure}[t]{0.99\textwidth}
         \centering
         \includegraphics[width=\textwidth]{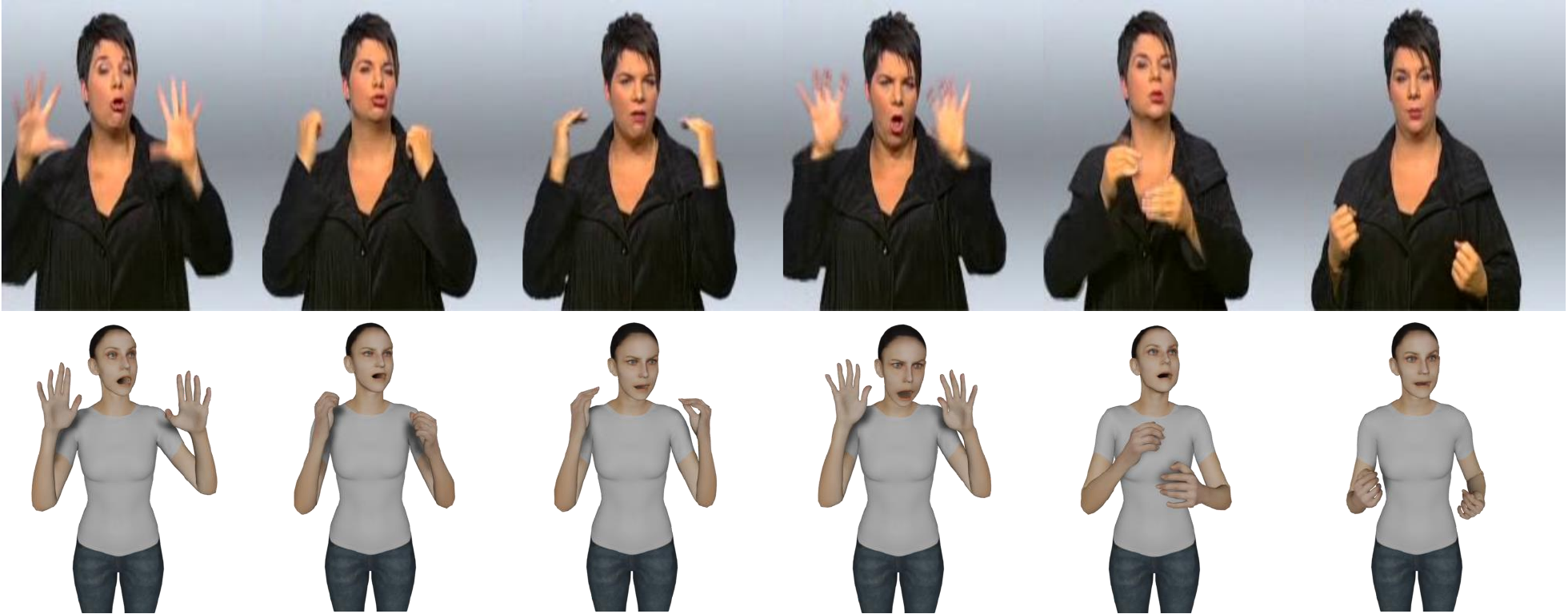}
         \caption{``At night it is mostly cloudy. At first there is only light rain locally, but later it starts to rain heavier in the south.''}
         \label{fig:supp_phoenix_B}
     \end{subfigure}
     \hfill
     \begin{subfigure}[t]{0.99\textwidth}
         \centering
         \includegraphics[width=\textwidth]{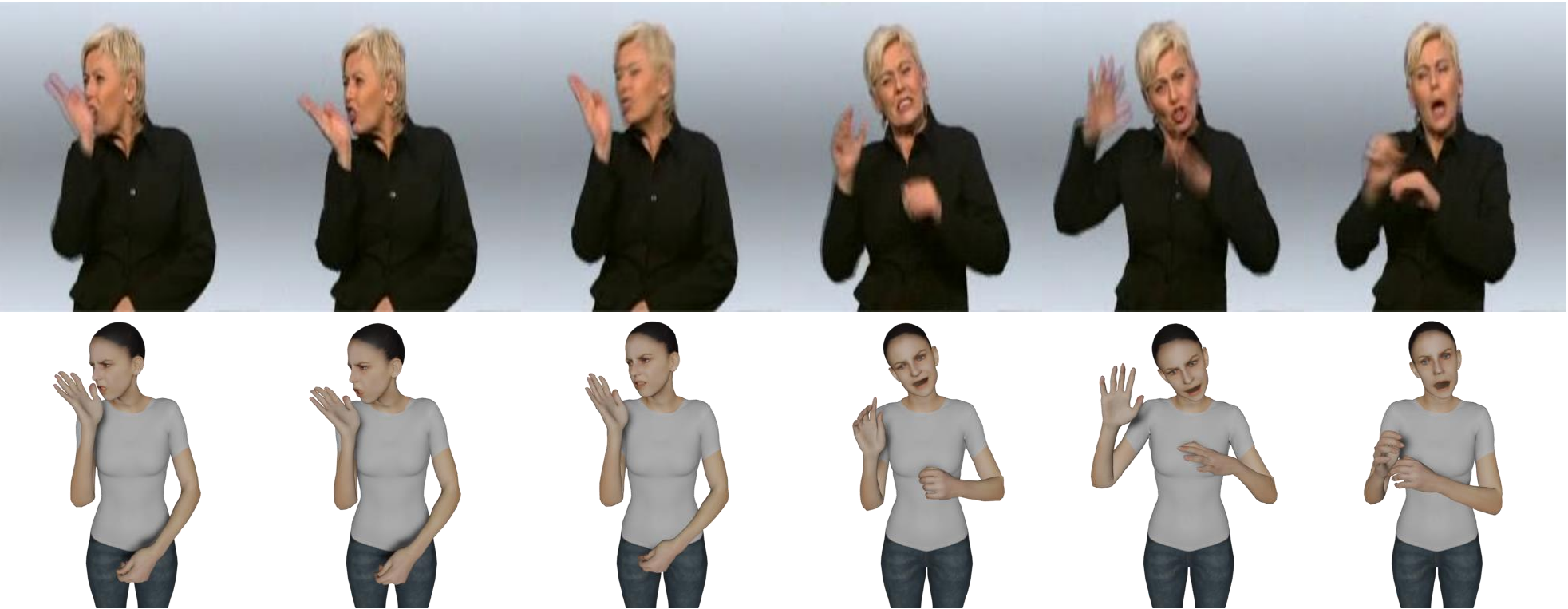}
         \caption{``There are still a few showers in the northeast, but they subside quickly during the night.''}
         \label{fig:supp_phoenix_C}
     \end{subfigure}
     \hfill
    \caption{Qualitative results on Phoenix-2014T~\cite{2014T}. In each sub-figure, we display the text in the caption, and show the ground truth sign video and our translation result in the first row and second row, respectively. We translate German into English.}
    \label{fig:supp_phoenix}
    \vspace{5mm}
\end{figure}

\begin{figure}[!ht]
     \centering
     \begin{subfigure}[t]{0.99\textwidth}
         \centering
         \includegraphics[width=\textwidth]{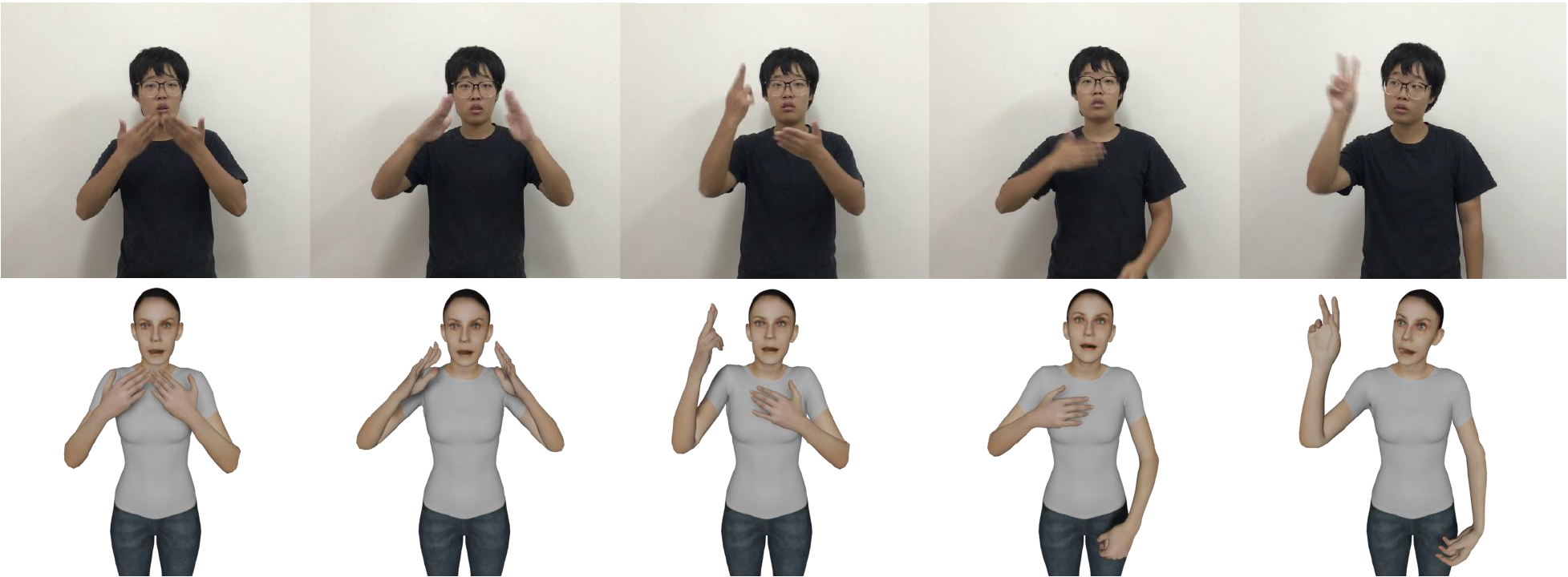}
         \caption{``My mother was sick, and the school allowed me to take leave to go home and visit her.''}
         \label{fig:supp_csl_A}
     \end{subfigure}
     \hfill
    \begin{subfigure}[t]{0.99\textwidth}
         \centering
         \includegraphics[width=\textwidth]{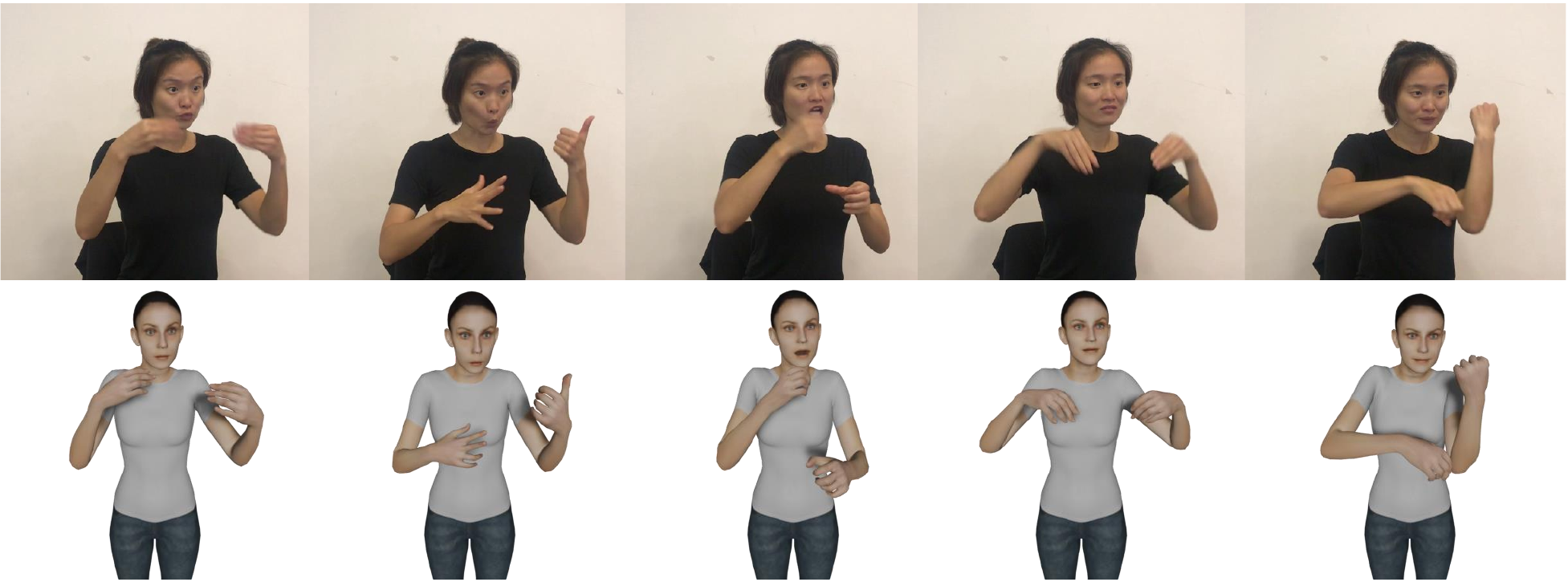}
         \caption{``My country's education system includes elementary education, vocational education, etc.''}
         \label{fig:supp_csl_B}
     \end{subfigure}
     \hfill
     \begin{subfigure}[t]{0.99\textwidth}
         \centering
         \includegraphics[width=\textwidth]{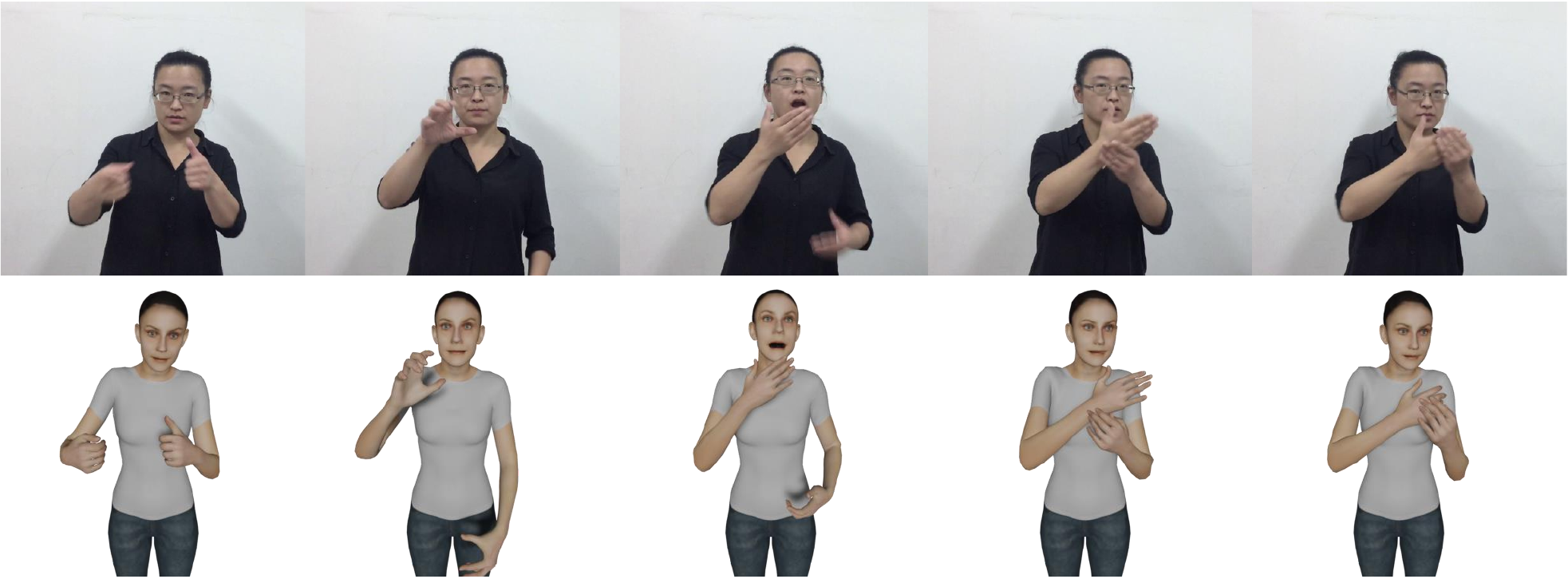}
         \caption{``Put the tea and cups away first, and then take out the watermelon from the refrigerator.''}
         \label{fig:supp_csl_C}
     \end{subfigure}
     \hfill

    \caption{Qualitative results on CSL-Daily~\cite{zhou2021improving}. In each sub-figure, we display the text in the caption, and show the ground truth sign video and our translation result in the first row and second row, respectively. We translate Chinese into English.}
    \label{fig:supp_csl}
\end{figure}

\noindent\textbf{Translation Results.} Please refer to the video demos in the supplementary material for additional visualizations. These demos display the ground truth sign videos alongside the baseline, SMPLify-X \cite{pavlakos2019expressive}, and our translation results, which are presented through a sign avatar. We also show several visualization results in Figure~\ref{fig:supp_phoenix} and Figure~\ref{fig:supp_csl} for Phoenix-2014T and CSL-Daily, respectively.

\begin{figure}[t]
     \centering
     \begin{subfigure}[t]{0.7\textwidth}
         \centering
         \includegraphics[width=\textwidth]{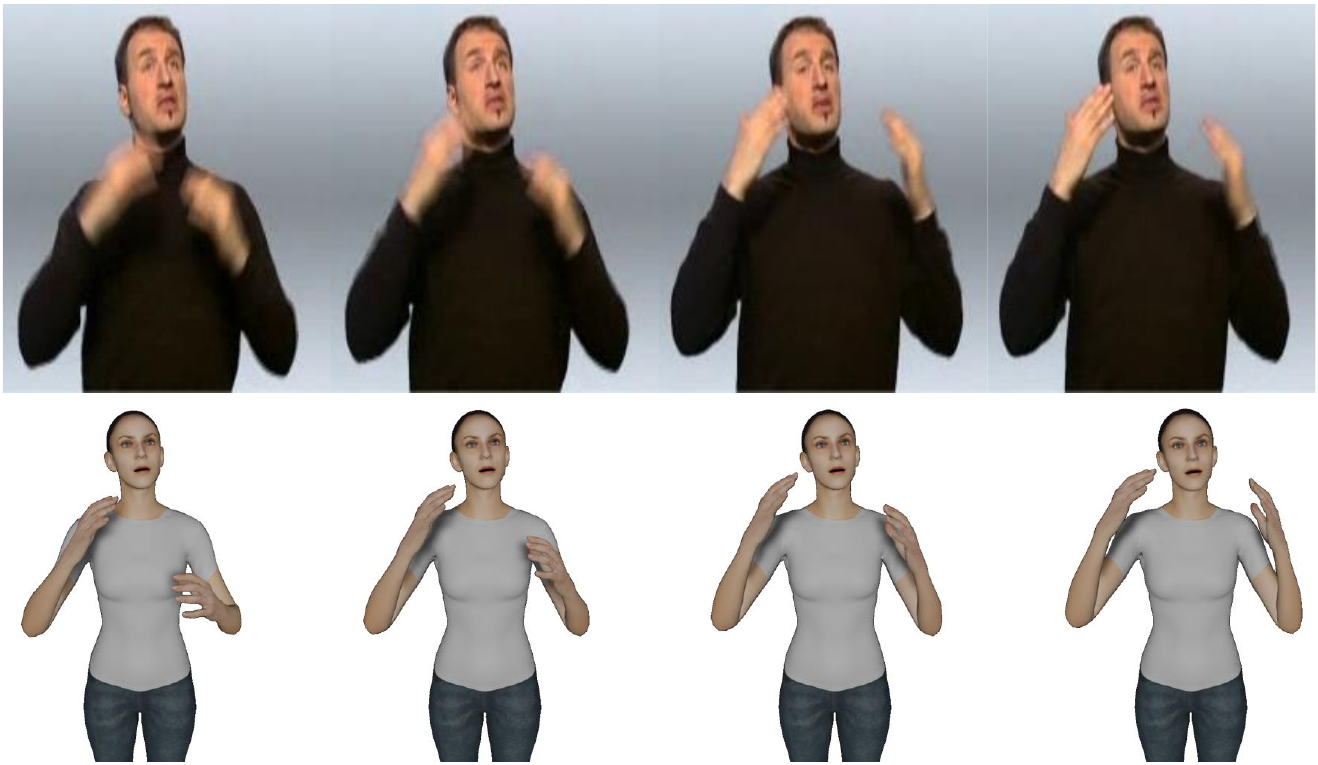}
         \caption{Example (a).}
         \label{fig:sign_connector_1}
     \end{subfigure}
     \hfill
    \begin{subfigure}[t]{0.7\textwidth}
         \centering
         \includegraphics[width=\textwidth]{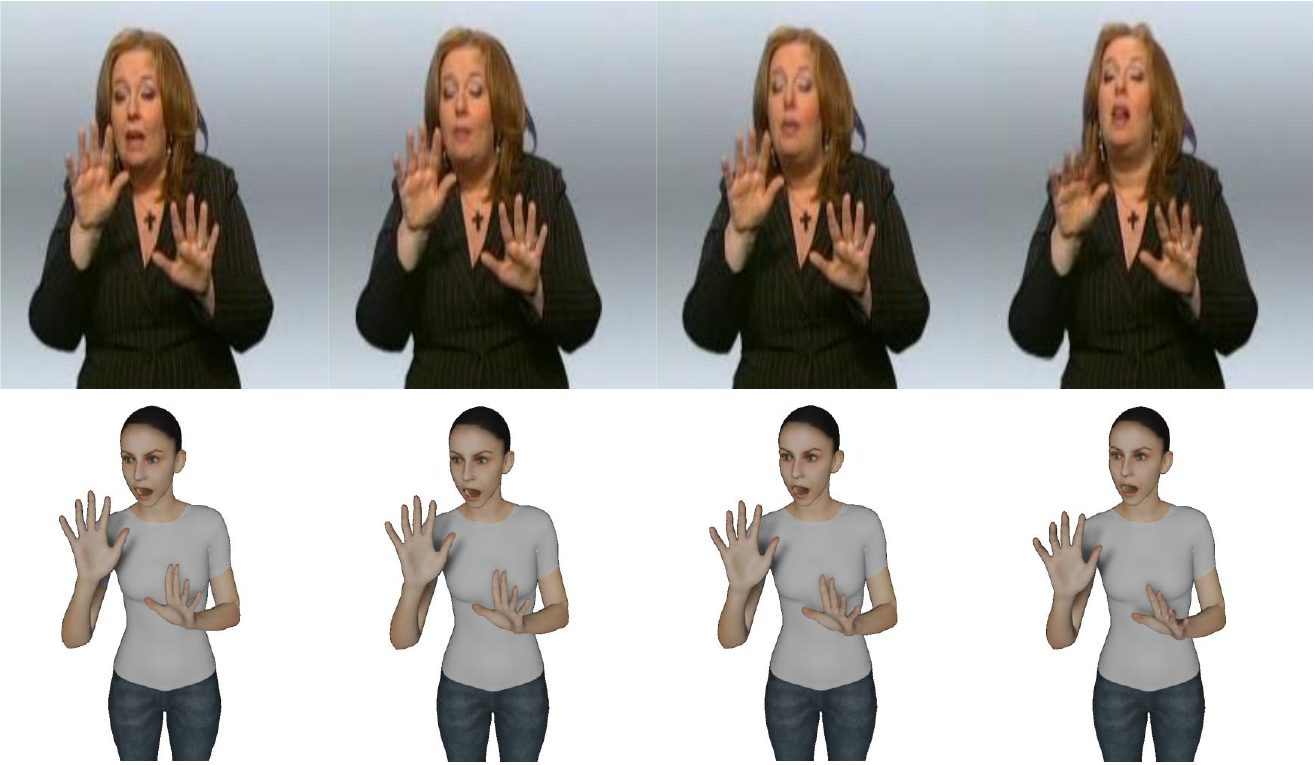}
         \caption{Example (b).}
         \label{fig:sign_connector_2}
     \end{subfigure}
     \hfill
     \begin{subfigure}[t]{0.7\textwidth}
         \centering
         \includegraphics[width=\textwidth]{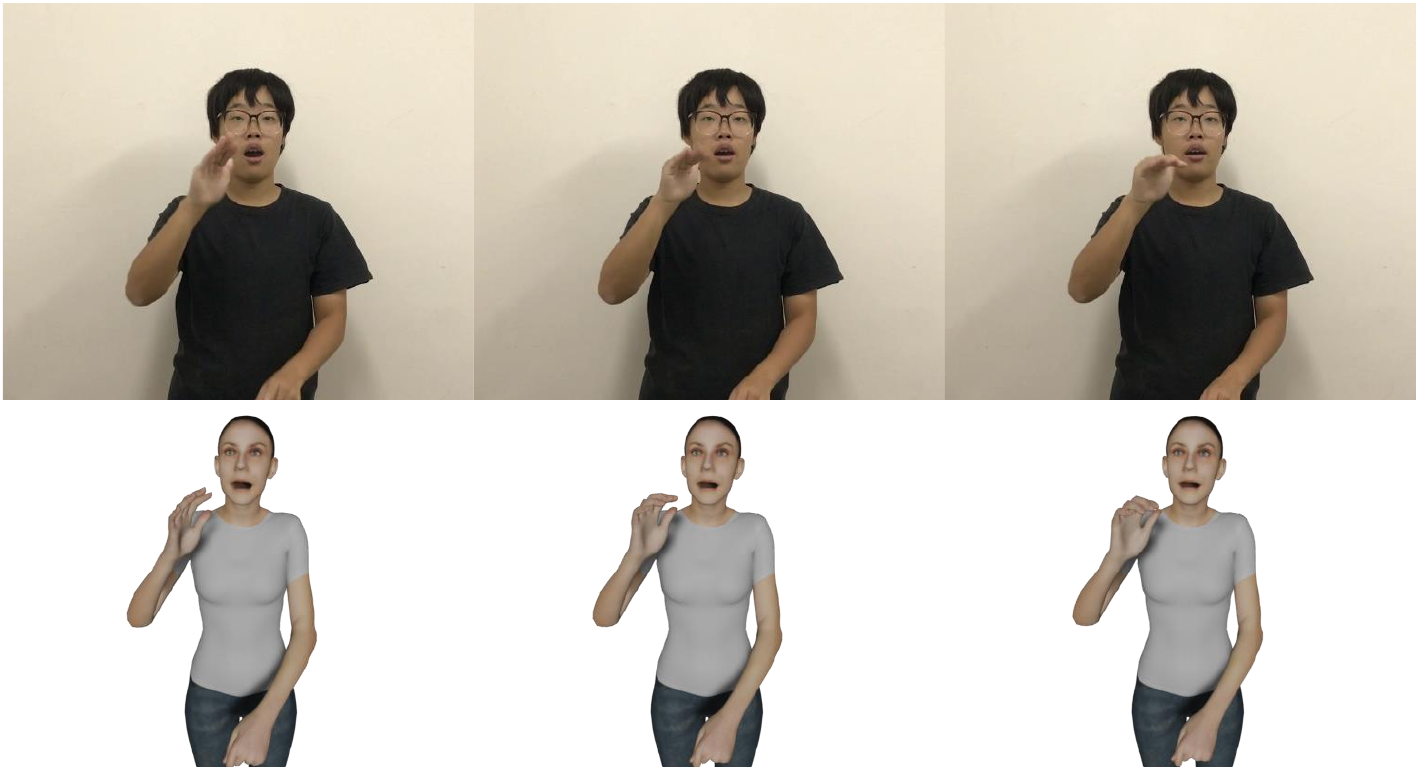}
         \caption{Example (c).}
         \label{fig:sign_connector_3}
     \end{subfigure}
     \hfill

    \caption{Qualitative comparison between the co-articulations generated by our sign connector and the corresponding ground truth. Three examples are randomly selected from Phoenix-2014T~\cite{2014T} (a and b) and CSL-Daily \cite{zhou2021improving} (c). In each sub-figure, the first row represents the ground truth, and the second row denotes the prediction.}
    \label{fig:vis_sign_connector}
    \vspace{5mm}
\end{figure}

\noindent\textbf{Sign Connector.} The objective of the sign connector is to predict the length (denoted as $L$) of co-articulations and to mimic these co-articulations by evenly interpolating $L$ frames between two adjacent signs in the 3D space. We present three generated co-articulations alongside their corresponding ground truths in Figure~\ref{fig:vis_sign_connector}.

\begin{figure}[t]
    \centering
    \includegraphics[width=0.85\textwidth]{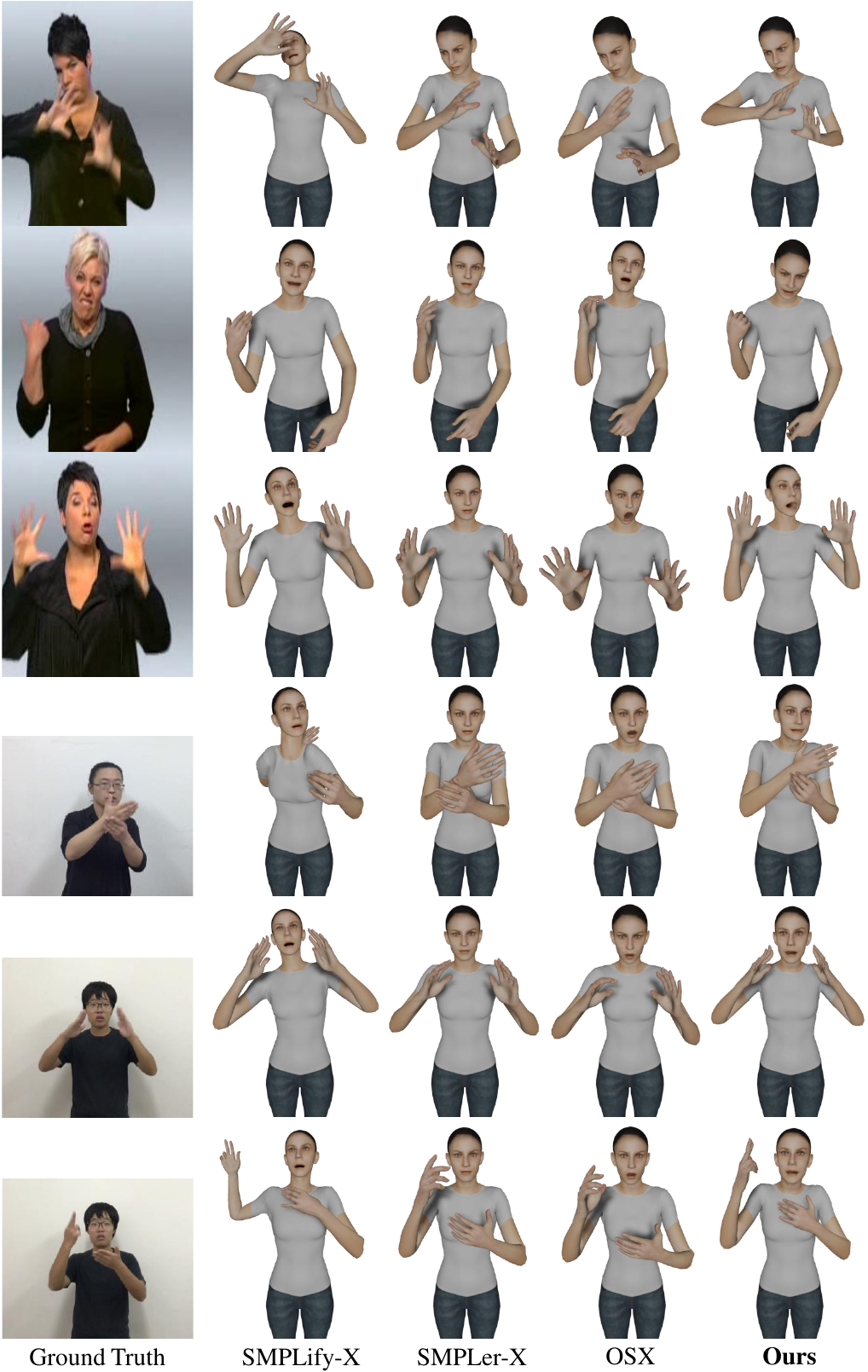}
    \caption{Qualitative comparison with other 3D sign estimation methods including SMPLify-X \cite{pavlakos2019expressive}, SMPLer-X \cite{cai2023smpler}, and OSX \cite{lin2023osx}, on Phoenix-2014T \cite{2014T} (the first three rows) and CSL-Daily \cite{zhou2021improving} (the last three rows).}
    \label{fig:supp_baseline}
\end{figure}

\noindent\textbf{Comparison with Other 3D Sign Estimators.}
In the main paper, we present SMPLSign-X, an enhancement of the original SMPLify-X \cite{pavlakos2019expressive}, tailored for 3D sign language estimation. In Table 2 of the main paper, we conduct a quantitative comparison with other state-of-the-art 3D sign estimation methods: SMPLify-X \cite{pavlakos2019expressive}, SMPLer-X \cite{cai2023smpler}, and OSX \cite{lin2023osx}. Additionally, we further demonstrate qualitative comparisons with these methods as shown in Figure \ref{fig:supp_baseline}. It is evident that our SMPLSign-X produces more visually appealing estimation results compared to all other methods.

\begin{figure}[t]
    \begin{minipage}[t]{.47\linewidth}
    \centering
    \includegraphics[width=0.99\linewidth]{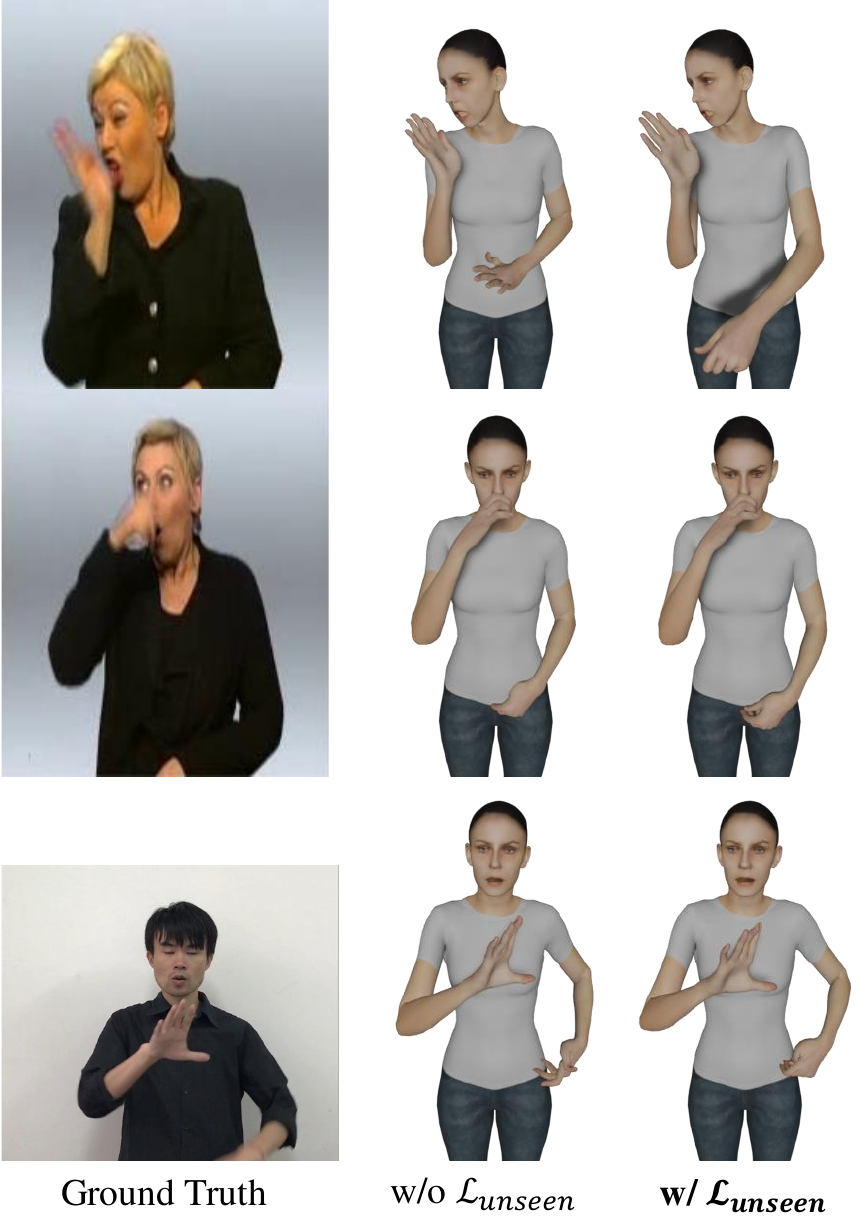}
    \captionof{figure}{Qualitative ablation study on $\mathcal{L}_{unseen}$ using Phoenix-2014T (the first two rows) and CSL-Daily (the last row).}
    \label{fig:supp_loss_unseen}
    \end{minipage}
    \hfill
    \begin{minipage}[t]{.49\linewidth}
    \centering
    \includegraphics[width=0.99\linewidth]{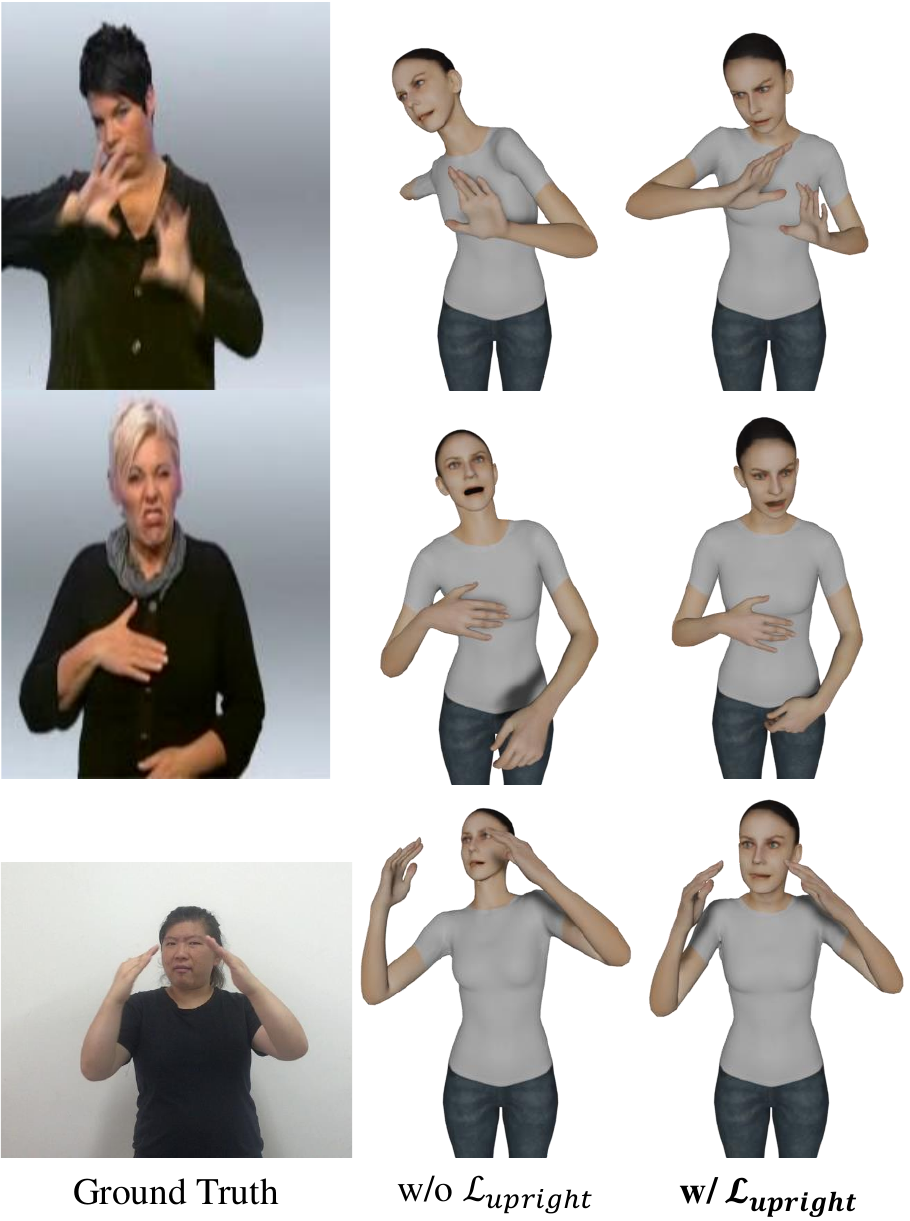}
    \captionof{figure}{Qualitative ablation study on $\mathcal{L}_{upright}$ using Phoenix-2014T (the first two rows) and CSL-Daily (the last row).}
    \label{fig:supp_loss_upright}
    \end{minipage}
\vspace{-5mm}
\end{figure}

\begin{figure}[t]
     \centering
     \includegraphics[width=0.99\linewidth]{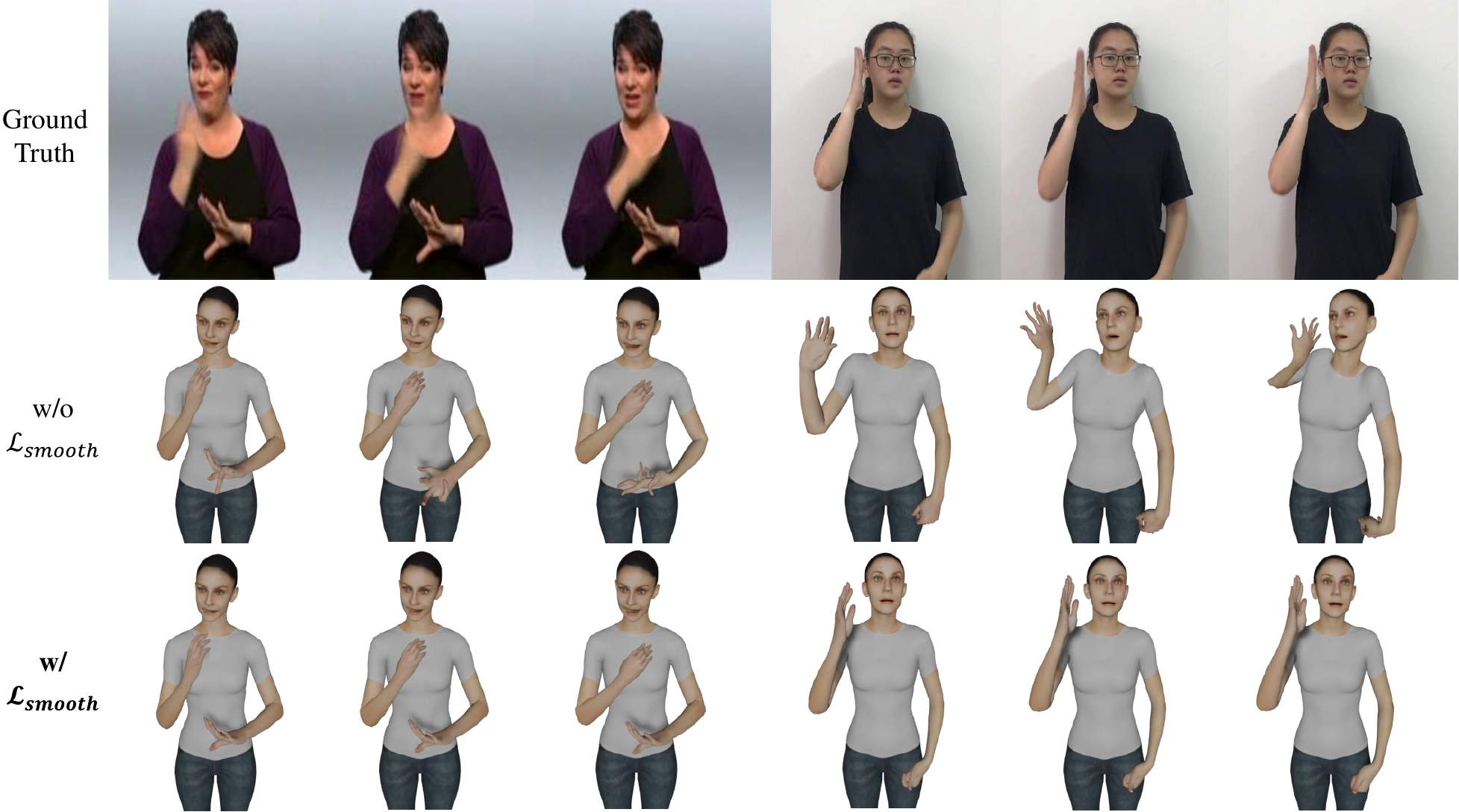}
     \caption{Qualitative ablation study on $\mathcal{L}_{smooth}$. We randomly sample three \textit{consecutive} frames from Phoenix-2014T \cite{2014T} (the first three columns) and CSL-Daily \cite{zhou2021improving} (the last three columns), respectively.}
     \label{fig:supp_loss_smooth}
\end{figure}

\noindent\textbf{Qualitative Ablation Study on Loss Functions.}
In Section 3.2 of the main paper, we introduce three loss functions to facilitate 3D sign estimation: $\mathcal{L}_{unseen}$ (Eq. 4), $\mathcal{L}_{upright}$ (Eq. 5), and $\mathcal{L}_{smooth}$ (Eq. 6). $\mathcal{L}_{unseen}$ is designed to draw the unseen keypoints closer to those of the rest pose; $\mathcal{L}_{upright}$ aims to encourage an upright posture in the upper body; and $\mathcal{L}_{smooth}$ preserves temporal consistency for generating visually appealing results. We have conducted a quantitative ablation study for these loss functions as shown in Table 4 in the main paper. Here we conduct a qualitative ablation study to further validate their effectiveness: as depicted in Figures \ref{fig:supp_loss_unseen}, \ref{fig:supp_loss_upright}, and \ref{fig:supp_loss_smooth}, it is evident that excluding any loss function degrades the estimation results.

\section{Broader Impacts and Limitations}
\noindent\textbf{Broader Impacts.} 
For the first time, we introduce a practical system for translating spoken language into sign language, with the translation results presented through a 3D avatar. The Spoken2Sign task is orthogonal and complementary to many sign language understanding tasks, such as sign language recognition and translation. Therefore, our system can further bridge the communication gap between the deaf and the hearing.

\noindent\textbf{Limitations.}
Although we have established a promising baseline for translating spoken languages to sign languages, several limitations still impact the translation results. First, understanding sign languages significantly suffers from the issue of data scarcity. Training on insufficient text-gloss sequence pairs may lead to sub-optimal Text2Gloss models, highlighting the critical need for large-scale sign language datasets. Second, our 3D sign estimator considers 2D keypoints estimated by HRNet~\cite{wang2020deep} as pseudo ground truths. However, inaccurate estimations can result in inferior 3D signs. A 2D keypoint estimator tailored for the sign language field might mitigate this issue. Lastly, as discussed in SMPLify-X~\cite{pavlakos2019expressive}, accurately estimating 3D skeletons with precise depth from a 2D image remains an unresolved research challenge. Introducing prior knowledge of sign languages could help eliminate depth ambiguity. We leave the resolution of these issues to future research.
\end{document}